\documentclass[10pt]{article}
\usepackage{ueprint}
\usepackage{xspace}
\usepackage{amsthm}
\usepackage{algorithm}
\usepackage[noend]{algpseudocode}
\newcommand{\calls}[1]{\hfill{\color{black!40}\footnotesize #1}}   %
\newcommand{\method}{\textsc{UltraEdit}\xspace}  
\newcommand{\sprite}{\textsc{Sprite}\xspace}%
\theoremstyle{plain}
\newtheorem{proposition}{Proposition}
\newtheorem{corollary}{Corollary}
\newtheorem{lemma}{Lemma}
\theoremstyle{remark}

\AtEndPreamble{\crefname{corollary}{Cor.}{Cors.}\crefname{proposition}{Prop.}{Props.}\Crefname{proposition}{Proposition}{Propositions}}
\newcommand{\R}{\mathbb{R}}
\newcommand{\E}{\mathbb{E}}
\newcommand{\Normal}{\mathcal{N}}
\newcommand{\src}{\mathrm{src}}
\newcommand{\tar}{\mathrm{tar}}
\newcommand{\Xsrc}{X^{\src}}
\newcommand{\Xtar}{X^{\tar}}
\newcommand{\csrc}{c_{\src}}
\newcommand{\ctar}{c_{\tar}}
\newcommand{\psrc}{p_{\src}}
\newcommand{\ptar}{p_{\tar}}
\newcommand{\tnext}{t'}
\newcommand{\Preach}{P_{\mathrm{reach}}}
\newcommand{\sig}{\mathrm{sig}}
\usepackage{colortbl}
\usepackage{tabularx}%
\definecolor{oursrow}{RGB}{231,241,251}
\definecolor{oursvarrow}{RGB}{243,247,253}
\newcommand{\ours}{\rowcolor{oursrow}}
\newcommand{\oursvar}{\rowcolor{oursvarrow}}
\newcommand{\un}[1]{\textsuperscript{\,#1}}%
\usepackage{tikz}
\definecolor{natureblue}{HTML}{4169C8}

\renewcommand{\paragraph}[1]{\vspace{.5em}\noindent\uerunin{#1}}
\usepackage[breaklinks,colorlinks]{hyperref}

\title{\method: Text-Guided Editing with Adaptive Coupling}
\subtitle{Text\uehyph Guided Editing with Adaptive Coupling}
\shorttitle{Text\uehyph Guided Editing with Adaptive Coupling}
\authors{%
  \href{https://oshikaka.github.io/}{Xinru Jiang}\textsuperscript{1}\quad
  Armand Comas\textsuperscript{2}\quad
  Lei Lu\textsuperscript{3}\quad
  Yizhou Zhao\textsuperscript{4}\quad
  Xi Xiao\textsuperscript{5}\\[0.25em]
  Chenrui Ma\textsuperscript{6}\quad
  Xiaomeng Yang\textsuperscript{1}\quad
  Xiao Wang\textsuperscript{7}\quad
  Octavia Camps\textsuperscript{1}\quad
  \href{https://lin-zhao-resolve.github.io/}{Lin Zhao}\textsuperscript{1,\dag}}
\affiliations{%
  \textsuperscript{1}Northeastern University\quad
  \textsuperscript{2}Google\quad
  \textsuperscript{3}Amazon\quad
  \textsuperscript{4}Carnegie Mellon University\\
  \textsuperscript{5}University of Alabama at Birmingham\quad
  \textsuperscript{6}University of Virginia\quad
  \textsuperscript{7}Oak Ridge National Laboratory}
\correspondence{\textsuperscript{\dag}Corresponding author.\quad E-mail:
  \href{mailto:jiang.xinru@northeastern.edu}{\texttt{jiang.xinru@northeastern.edu}},
  \href{mailto:zhao.lin1@northeastern.edu}{\texttt{zhao.lin1@northeastern.edu}}}
\links{%
  \preprintlink[primary]{globe}{https://oshikaka.github.io/UltraEdit/}{Project Page}%
  \preprintlink[primary]{github}{https://github.com/Oshikaka/UltraEdit}{Code}%
  \preprintlink[primary]{database}{https://huggingface.co/datasets/Oshikaka/SPRITE}{SPRITE data}}

\begin{document}
\begin{abstract}

Making substantial structural changes while preserving
subject identity and surrounding content remains
challenging in text-guided editing.
We analyze this trade-off through two aspects of flow-based
editing: the editing start time, and the source reference
path with its associated target update.
Through theoretical analysis and controlled experiments,
we examine how these choices affect structural editing
and source preservation, and how their effects differ
across regions.
These findings motivate \method, a training-free,
inversion-free editor combining spatially adaptive coupling
with source anchoring.
Inside an attention-selected region, we initialize the
source and target inputs as pure noise and use an SDE
reference path with a matching posterior update.
Outside, we start later and use fresh reference noise
with Euler updates.
On \sprite, our benchmark of 491 mostly structural edits,
\method achieves 78.0\% edit success, compared with
50.1\% for FlowEdit.
It also achieves 53.05\% video editing accuracy on
FiVE-Bench, exceeding the results reported by the
benchmark for other editors.
Beyond images and video, we present that the same
update rule enables audio, speech, and 3D editing without
retraining.
\end{abstract}

\maketitle
\begin{center}
\begin{minipage}{\textwidth}%
\vspace{-0.6em}
{\centering\includegraphics[width=0.86\textwidth]{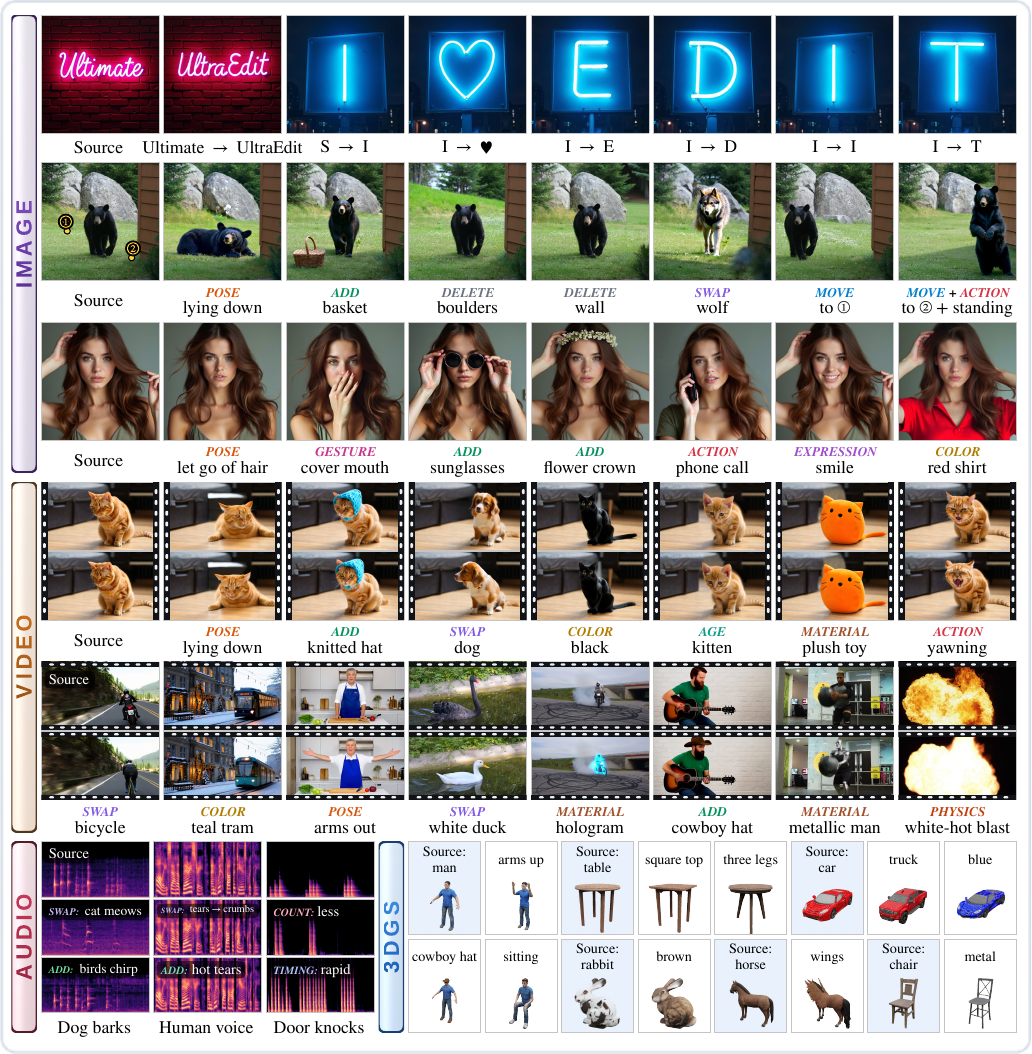}\par}
\captionof{figure}{\textbf{The same update supports diverse edits across multiple modalities.}
Without training, \method edits images and videos to
change shape, pose, motion, object count, color, and material. It also adds, removes, swaps, and moves objects to specified locations, rewrites scene text, and supports simultaneous or sequential edits while preserving unrelated content.
Beyond image and video editing, the same update extends
to audio, including sound and speech, and 3D assets
generated by TRELLIS.}
\label{fig:teaser}
\end{minipage}
\vspace{0.6em}
\end{center}

\makeabstract
\suppressfloats[t]
\section{Introduction}
\label{sec:intro}

Text-guided editing aims to make the changes specified
by a prompt while preserving unrelated content.
Inversion-based methods map the source image to noise
and then generate an edited image under the target prompt.
However, the resulting edit can also alter details
that should remain unchanged.
To improve source preservation,
some methods~\cite{kulikov2025flowedit,yoon2025splitflow,lu2026chordedit}
use directly noised source states to guide editing,
avoiding the need for an inversion trajectory.

\begin{figure*}[t]
\centering
\includegraphics[width=\textwidth]{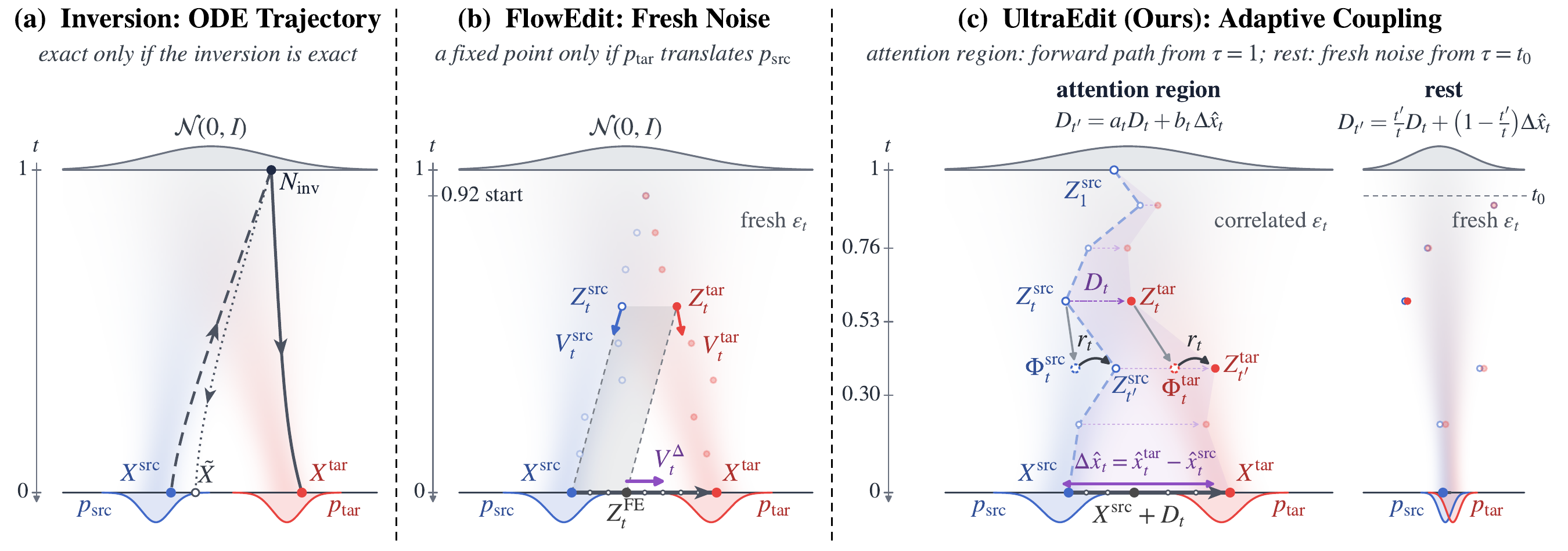}
\caption{\textbf{Source--target coupling across editing methods.}
(a) Inversion obtains a source ODE path before target sampling; $\tilde X$ regenerates from $N_{\mathrm{inv}}$ with the source prompt. (b) FlowEdit constructs fresh noisy source references and uses a velocity-difference update. (c) \method assigns a correlated forward path and posterior update to an adaptive region, while using fresh references and later starts outside the attention region. One step $t\to t'$ is drawn in the form of \cref{lem:target}: light grey arrows are the model updates $\Phi_t$, ending at the dotted rings, and the same $r_t$ (dark arrows) moves the source ring to the forward-path state and the target ring to $Z^{\mathrm{tar}}_{t'}$.}
\label{fig:coupling}
\end{figure*}

Yet, these methods may also retain source structures
that the target prompt asks to change.
As shown in \cref{fig:paths}, an inversion-based editor
opens the peacock's tail but alters unrelated details,
whereas inversion-free FlowEdit~\cite{kulikov2025flowedit} preserves the scene but leaves the tail closed.
This raises the question:
\textit{How can inversion-free editors enable substantial
structural changes while preserving subject identity
and surrounding content?}

We investigate two choices that govern source influence: the editing start time, which controls how much source information remains at initialization, and the source reference path with its associated target update, which determines how the source guides subsequent steps.
Our theoretical analysis in a simplified model shows that starting later can make the requested structural
change less likely.
Controlled experiments likewise show that starting
earlier improves structural edit success.
Following FlowEdit's formulation, we show that Fresh Noise + Euler averages the editing direction over reference noise in the small-step limit.
Based on this observation, we study a forward SDE
reference path with posterior updates matched to
its reverse conditional distribution.
Our experiments show that applying this configuration
globally improves edit success over Fresh Noise + Euler
but increases background changes.
Together, the findings motivate adapting start
time, reference path, and update rule to the different
needs of edited and surrounding regions, improving
the editing--preservation trade-off.

We propose \textbf{\method}, a training-free,
inversion-free editor combining adaptive source--target
coupling with source anchoring for content preservation.
The method first uses attention to identify the editing
region, then progressively updates the edit using paired
source and target predictions.
In the initial editing region, paired inputs start from
pure noise and editing uses SDE + Posterior; surrounding
regions start later with Fresh Noise + Euler.
During sampling, target attention guides the expansion
of the editing region to accommodate new geometry.

We also introduce \textbf{\sprite}, a benchmark focused
on structural editing.
Evaluations on images and videos show that \method
improves edit success and source preservation over
prior editors.
Beyond images and video, we present that the same
update enables audio, speech, and 3D editing without
retraining the base model. 

\section{Related Work}
\label{sec:related}
Existing text-guided editors use source information
in different ways to balance editing and preservation.
FlowEdit~\cite{kulikov2025flowedit} independently
resamples noisy source references at each step,
while CycleDiffusion~\cite{wu2023cyclediffusion}
and Sync-SDE~\cite{zhang2026syncsde} couple target
sampling to stochastic source trajectories.
Regional methods regulate where and how strongly
changes are applied.
Differential Diffusion~\cite{levin2025differential}
controls editing strength through spatially varying
denoising schedules, while
DiffEdit~\cite{couairon2023diffedit} localizes changes
using masks inferred from prompt-conditioned predictions.
To preserve source appearance during generation,
FreeFlux~\cite{wei2025freeflux} and
SynPS~\cite{chen2026synps} let target queries access
source attention keys and values.
SpotEdit~\cite{qin2026spotedit} identifies regions
whose predicted content remains perceptually similar
to the source and preserves them by reusing source
features.
We analyze how the start time, reference path, and
update rule affect structural editing and preservation,
motivating spatially adaptive source--target coupling.

\section{Preliminaries}
\label{sec:prelim}

\paragraph{Flow matching.}
We use $t=0$ for clean data and $t=1$ for pure Gaussian noise. We consider flow matching~\cite{lipman2023flow} with linear conditional paths, as used in rectified flow~\cite{liu2023rectified}:
\begin{equation}
z_t=(1-t)x+t\varepsilon,\qquad \varepsilon\sim\Normal(0,I),
\label{eq:prelim-interpolation}
\end{equation}
where $x$ is a clean latent. 
Given a text condition $c$, a velocity field $v(z,t,c)$ is learned by regressing the interpolation velocity $\varepsilon-x$.
Generation integrates $dz_t=v(z_t,t,c)\,dt$ from $t=1$ to $0$. 
The posterior-mean estimate of the clean latent is $\hat x(z,t,c)=z-t\,v(z,t,c)$.

\paragraph{Editing by inversion.}
As illustrated in \cref{fig:coupling}(a), let $Z_t^\src$ and $Z_t^\tar$ denote latent trajectories under the source and target prompts $\csrc$ and $\ctar$. Inversion integrates the source ODE from $Z_0^\src=\Xsrc$ at $t=0$ to the noise $Z_1^\src=N_{\mathrm{inv}}$ at $t=1$. Target sampling then runs from $Z_1^\tar=N_{\mathrm{inv}}$ to the edited latent $Z_0^\tar=\Xtar$ at $t=0$.

\paragraph{Inversion-free editing.}
FlowEdit~\cite{kulikov2025flowedit} writes inversion-based editing through the displacement $D_t=Z_t^\tar-Z_t^\src$:
\begin{equation}
\frac{dD_t}{dt}=V_t^\tar-V_t^\src,
\label{eq:prelim-inversion}
\end{equation}
where $V_t^\src=v(Z_t^\src,t,\csrc)$ and $V_t^\tar=v(Z_t^\tar,t,\ctar)$. The path $\Xsrc+D_t$ connects $\Xsrc$ at $t=1$ to $\Xtar$ at $t=0$, but still requires the inverted source trajectory.

To avoid inversion, FlowEdit instead evolves a \emph{direct editing state} $Z_t^{\mathrm{FE}}$, initialized at $Z_{t_0}^{\mathrm{FE}}=\Xsrc$ at its start time $t_0$ and ending at the edited output $Z_0^{\mathrm{FE}}=\Xtar$. At each step, it constructs two noisy model inputs:
\begin{equation}
\begin{aligned}
Z_t^\src&=(1-t)\Xsrc+t\varepsilon_t,\quad \varepsilon_t\sim\Normal(0,I),\\
Z_t^\tar&=Z_t^{\mathrm{FE}}+Z_t^\src-\Xsrc.
\end{aligned}
\label{eq:prelim-flowedit-copies}
\end{equation}
The corresponding displacement is
$D_t=Z_t^{\mathrm{FE}}-\Xsrc$ and
\begin{equation}
\frac{dZ_t^{\mathrm{FE}}}{dt}=\E_{\varepsilon_t}\!\left[V_t^\tar-V_t^\src\right]
\label{eq:prelim-flowedit}
\end{equation}
is integrated toward $t=0$.
At each step, FlowEdit draws a fresh $\varepsilon_t$, independent of previous steps, to estimate expected velocity difference, then updates $Z_t^{\mathrm{FE}}$, as in \cref{fig:coupling}(b).

\section{Analyzing Flow-Based Editing}
\label{sec:analysis}
\def\uePathsW{0.36}
\newsavebox{\uePathsBox}
\sbox{\uePathsBox}{\begin{minipage}{\uePathsW\linewidth}\centering
\includegraphics[width=\linewidth]{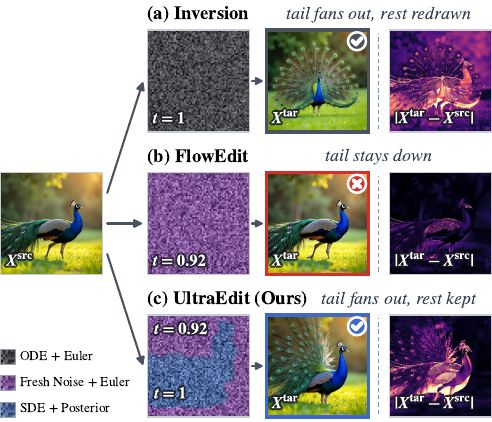}
\end{minipage}}
\newcommand{\uePathsCaption}{\captionof{figure}{\textbf{Editing methods differ in how they balance structural change and source preservation.} The prompt asks the peacock to fan its tail.
(a) Inversion opens the tail but alters unrelated details.
(b) FlowEdit preserves the scene but leaves the tail closed.
(c) \method opens the tail while better preserving
the scene.
}
\label{fig:paths}}
\newsavebox{\uePathsText}
\sbox{\uePathsText}{\begin{minipage}{\dimexpr0.96\linewidth-\uePathsW\linewidth\relax}
\cref{fig:paths} motivates the question: 

\par\smallskip\noindent
\begingroup
\definecolor{questionpurple}{HTML}{6A3B92}%
\begin{tikzpicture}
\node[fill=questionpurple!5, inner xsep=6pt, inner ysep=5pt,
      text width=\dimexpr\linewidth-12pt\relax, align=left] (editingquestion) {%
  \textcolor{questionpurple}{\textbf{Question:}} Why do existing inversion-free editors struggle to make substantial structural changes while preserving identity and surrounding content?};
\draw[questionpurple, line width=2pt]
  (editingquestion.north west) -- (editingquestion.south west);
\end{tikzpicture}
\endgroup
\par\smallskip\noindent
To tackle this question, we examine two aspects of source-target coupling: the editing \emph{start time} and the \emph{source reference path and its associated target update}. 
The former determines how much source information remains in the initial noisy state, while the latter governs how the source image influences subsequent target updates.
Through theoretical analysis and experiments, we examine how these choices affect structural change and source preservation.
\end{minipage}}
\typeout{uePaths: figure \the\dimexpr\ht\uePathsBox+\dp\uePathsBox\relax, text \the\dimexpr\ht\uePathsText+\dp\uePathsText\relax}
\par\smallskip\noindent\begin{minipage}{\linewidth}
\raisebox{-0.5\height}{\usebox{\uePathsBox}}\hfill\raisebox{-0.5\height}{\usebox{\uePathsText}}\par
\uePathsCaption
\end{minipage}\par\medskip

\subsection{Theoretical Analysis}
\label{sec:analysis-dynamics}

\paragraph{Separating initialization from subsequent updates.}
For an edit initialized at time $\tau$, we set $Z_\tau^\tar=Z_\tau^\src$, i.e., $D_\tau=0$.
For a step from $t$ to $\tnext<t$, FlowEdit uses Euler's
method to update the displacement $D_t$ as follows:
\[D_{\tnext}
=D_t-(t-\tnext)(V_t^\tar-V_t^\src)
=\frac{\tnext}{t}D_t+\left(1-\frac{\tnext}{t}\right)\Delta\hat x_t,
\]
where $\Delta\hat x_t=\hat x(Z_t^\tar,t,\ctar)-\hat x(Z_t^\src,t,\csrc)$.
To include other update rules, we consider the more
general affine form with coefficients
$a_t$ and $b_t$:
\begin{equation}
D_{\tnext}=a_tD_t+b_t\Delta\hat x_t,
\qquad 0\le\tnext<t\le1.
\label{eq:offset}
\end{equation}
We rewrite \cref{eq:offset} as a target state update to identify how the source influences subsequent steps. 

\begin{lemma}[Target dynamics]\label{lem:target}
Let $\Phi_t^c(Z)=a_tZ+b_t\hat x(Z,t,c)$. Then \cref{eq:offset} is equivalent to:
\begin{equation}
Z_{\tnext}^\tar=\Phi_t^\tar(Z_t^\tar)+r_t, \ \ 
r_t=Z_{\tnext}^\src-\Phi_t^\src(Z_t^\src).
\label{eq:target}
\end{equation}
\end{lemma}
The proof is given in Supp.~\ref{supp:proofs-target}.
This representation allows us to
analyze subsequent source influence separately from initialization.
The start time determines the initial noisy state $Z_\tau^\src$.
The source reference and update rule influence subsequent target updates through $r_t$.
For ODE inversion, $r_t=0$ when the source transition matches
the source-conditioned update $\Phi_t^\src$\footnote{Finite-step
approximations may leave a numerical residual.}.
For inversion-free methods, subsequent
updates can still depend on the source image through $r_t$,
even when editing starts from pure noise.

\paragraph{Reference paths and target updates.}
In the small-step limit, FlowEdit follows the velocity
difference averaged over reference noise
(Supp.~\cref{prop:average}).
We construct an alternative source reference path using
a forward Gaussian SDE from $Z_0^\src=\Xsrc$.
To account for the dependence between successive reference
states, we derive a matching posterior update from the
path's reverse conditional distribution.
For $0\le\tnext<t\le1$, the forward transition is:
\begin{equation}
\begin{gathered}
Z_t^\src=kZ_{\tnext}^\src+sN_t,\qquad N_t\sim\Normal(0,I),\\
k=\frac{1-t}{1-\tnext},\qquad s^2=t^2-k^2\tnext^2.
\end{gathered}
\label{eq:forward-path}
\end{equation}
Although the $N_t$ are independent, each $Z_t^\src$ builds on the previous state, so the total noise $\varepsilon_t=(Z_t^\src-(1-t)\Xsrc)/t$ is correlated across steps, unlike the fresh $\varepsilon_t$ of \cref{eq:prelim-flowedit-copies}.
At each time $t$ along the forward path, $Z_t^\src\mid\Xsrc\sim\Normal((1-t)\Xsrc,t^2I)$, matching the marginal distribution of FlowEdit's source references.
Having specified the source reference path, we next derive a matching posterior update for the target. The path's reverse conditional is:
\begin{equation}
Z_{\tnext}^\src\mid(Z_t^\src,\Xsrc=x)\sim
\Normal(a_tZ_t^\src+b_tx,\varsigma^2I).
\label{eq:posterior-step}
\end{equation}
Our posterior update follows \cref{eq:offset}, with
\begin{equation}
a_t=\frac{k\tnext^2}{t^2},\qquad
b_t=\frac{(1-\tnext)s^2}{t^2},\qquad
\varsigma=\frac{s\tnext}{t}.
\label{eq:posterior-coefficients}
\end{equation}
This construction builds on posterior-based coupling and synchronized stochastic dynamics~\cite{wu2023cyclediffusion,nie2024blessing,zhang2026syncsde}. 
The theoretical analysis is provided in Supp.~\cref{prop:sync}.

\paragraph{How start time affects structural editing.}
\label{sec:when}
Having specified the reference path and update rule, we ask
whether a late start still limits structural change.
We analyze this construction in continuous time, with all
positions sharing a start time $0<\tau<1$.

\begin{proposition}[Effect of start time]\label{prop:reach}
Let $\psrc$ and $\ptar$ denote the source and target distributions, and assume exact models for both.
\emph{(i)} For $\Xsrc\sim\psrc$, the edited output, conditioned
on the noisy starting state $Z_\tau^\src$, follows the target
posterior $\ptar(x_0\mid z_\tau=Z_\tau^\src)$.
\emph{(ii)} For a simplified analysis, let
$\ptar=(1-\pi_{\mathrm{new}})\psrc
      +\pi_{\mathrm{new}}p_{\mathrm{new}}$,
where $\psrc$ and $p_{\mathrm{new}}$ are Gaussian distributions
with shared covariance, representing the original and modified
structures, and $\pi_{\mathrm{new}}$ is the
weight of the modified component.
Let $\Preach(\tau)$ denote the probability of achieving the
structural change when editing starts at $\tau$.
Later starts (smaller $\tau$) lower or leave this probability
unchanged. As $\tau\uparrow1$, it approaches $\pi_{\mathrm{new}}$.
\textup{The expression for $\Preach$ and proof are given in
Supp.~\ref{supp:when-reach}.}
\end{proposition}

\subsection{Experimental Analysis}
\label{sec:analysis-empirical}

We conduct controlled experiments on {\bf \sprite}, our benchmark focused on structural editing, using FLUX.1-dev.

\def\ueExitW{0.86}
\newsavebox{\ueExitL}\newsavebox{\ueExitR}
\begin{figure}[t]
\sbox{\ueExitL}{\begin{minipage}[t]{0.5\linewidth}\vspace{0pt}\centering\vspace*{-3.3pt}%
\includegraphics[width=\ueExitW\linewidth]{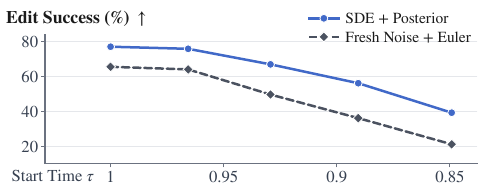}
\caption{{\textbf{Effect of start time on edit success.}}
Every token starts at $\tau$ under one configuration over the
whole image, without source anchoring (no shared keys and values or scaling-back step).
}
\label{fig:exitcurve}
\end{minipage}}%
\sbox{\ueExitR}{\begin{minipage}[t]{0.46\linewidth}\vspace{0pt}\centering
\captionof{table}{\textbf{Comparison of different global and regional configurations.}
The two global configurations start every token from $t=1$.
All configurations are evaluated without source anchoring, 16 steps.
We report edit success (Edit), background PSNR (BG), and background L1 error (BG L1).}
\label{tab:analysis-global}
\footnotesize
\setlength{\tabcolsep}{3pt}
\begin{tabular*}{\linewidth}{@{\extracolsep{\fill}}ll|rrr@{}}
\toprule
Reference & Update & Edit\un{\%} $\uparrow$ & BG\un{dB} $\uparrow$ & BG L1\un{0--255} $\downarrow$\\
\midrule
Fresh Noise & Euler &65.4&15.98&19.86\\
SDE & Posterior &76.9&13.63&30.84\\
Combined & Combined &77.7 &15.56&20.33\\
\bottomrule
\end{tabular*}
\end{minipage}}%
\typeout{ueExit: left \the\dimexpr\ht\ueExitL+\dp\ueExitL\relax, right \the\dimexpr\ht\ueExitR+\dp\ueExitR\relax}%
\noindent\usebox{\ueExitL}\hfill\usebox{\ueExitR}
\end{figure}

\paragraph{How does start time affect structural editing?}
In \cref{fig:exitcurve}, we vary the common start time
$\tau$ for all tokens while keeping other settings fixed
within each configuration.
Delaying the start from $\tau=1$ to $0.85$ reduces edit
success from 76.9\% to 39.1\% for SDE + Posterior and
from 65.4\% to 21.0\% for Fresh Noise + Euler.
Earlier starts improve edit success
for both configurations.

\paragraph{Can one configuration balance editing and preservation?}
Across all tested starts, SDE + Posterior achieves
11.5-20.0 percentage points higher edit success
than Fresh Noise + Euler, as in \cref{fig:exitcurve}.
However, as shown in \cref{tab:analysis-global}, when both configurations are applied globally
from $\tau=1$, SDE + Posterior
causes larger background changes, increasing BG L1
from 19.86 to 30.84.
This trade-off motivates different configurations
for the editing region and its surroundings.

\paragraph{Should configurations vary by region?}
Motivated by this trade-off, we combine SDE + Posterior
from $\tau=1$ inside the editing region with Fresh Noise
+ Euler from a later start outside.
As shown in \cref{tab:analysis-global}, this regional
configuration achieves 77.7\% edit success, compared
with 76.9\% for global SDE + Posterior, while reducing
BG L1 from 30.84 to 20.33, close to the 19.86 achieved
by global Fresh Noise + Euler.
These results suggest that assigning different
configurations to edited and surrounding regions
improves the editing--preservation trade-off.
An idealized analysis of this regional design is provided
in Supp.~\cref{prop:leak,cor:regional}.

\par\smallskip\noindent
\begingroup
\begin{tikzpicture}
\node[fill=natureblue!5, inner xsep=6pt, inner ysep=6pt,
      text width=\dimexpr\linewidth-12pt\relax, align=left] (editingtakeaways) {%
\textcolor{natureblue}{\textbf{Takeaways:}}
\begin{list}{\textbullet}{%
  \setlength{\leftmargin}{11pt}%
  \setlength{\labelsep}{4pt}%
  \setlength{\itemsep}{3pt}%
  \setlength{\topsep}{3pt}%
  \setlength{\parsep}{0pt}%
  \setlength{\partopsep}{0pt}%
}
\item
Earlier starts improve structural edit success.

\item Applying SDE + Posterior everywhere involves a trade-off
between structural editing and background preservation.

\item Using different configurations in different regions
improves this trade-off.
\end{list}};
\draw[natureblue, line width=2pt]
  (editingtakeaways.north west) -- (editingtakeaways.south west);
\end{tikzpicture}
\endgroup
\par\smallskip\noindent

\section{UltraEdit}
\label{sec:method}

\method turns the observations of \cref{sec:analysis} into two components. 
\emph{Spatially adaptive coupling} assigns each token
a start time, a reference path, and an update rule,
controlling when and how editing proceeds at each location.
\emph{Source anchoring} uses the paired source branch
to preserve content unrelated to the edit.
\cref{fig:coupling,alg:method} summarize the procedure.

\subsection{Spatially Adaptive Coupling}
\begin{algorithm}[b]%
\caption{\textbf{\method algorithm.} Grey: NFE per line.}
\label{alg:method}
\small
\begin{algorithmic}[1]
\Require $\Xsrc$, $\csrc$, $\ctar$, edited words;
$16$ steps; $t_0$, $t_r$
\State $\mu_0\gets$ attention probes at $t=0.7,0.5$;
$\mu\gets\mu_0$; $\tau\gets1$ on $\mu_0$,
$t_0$ off $\mu_0$ \calls{4}
\State $Z^{\src,\mathrm{fw}}\gets$ forward path of
$\Xsrc$, \cref{eq:forward-path} \calls{0}
\State $D\gets0$ \Comment{current displacement $D_t$}
\For{each step $t\to\tnext$}
  \State $Z_t^\src\gets
  \mu\odot Z_t^{\src,\mathrm{fw}}
  +(1-\mu)\odot[(1-t)\Xsrc+t\varepsilon_t]$
  \State $\hat x_t^\src,\hat x_t^\tar\gets
  \hat x(Z_t^\src,t,\csrc),\,
  \hat x(Z_t^\src+D,t,\ctar)$ \calls{2}
  \If{first step with $t\le t_r$}
    \State $\omega\gets$ agreement of
    $\hat x_t^\tar$ and $\hat x_t^\src$ on $\mu$
    \State $D\gets(1-\omega)\odot D$
    \Comment{scale back where the copies agree}
    \State recompute $\hat x_t^\src$, $\hat x_t^\tar$
    \calls{2}
  \EndIf
  \State on $\mu$:
  $D\gets a_tD+b_t(\hat x_t^\tar-\hat x_t^\src)$
  \State off $\mu$, $t\le\tau$:
  $D\gets\tfrac{\tnext}{t}D+
  (1-\tfrac{\tnext}{t})(\hat x_t^\tar-\hat x_t^\src)$
  \State \algorithmicif\ $\tnext>t_r$
  \algorithmicthen\ expand $\mu$;
  new tokens: $\tau\gets\max(\tau,\tnext)$ \calls{0}
\EndFor
\State \Return $\Xtar=\Xsrc+D$
\end{algorithmic}
\end{algorithm}%
\label{sec:method-coupling}

\paragraph{Regional reference and update.}
We implement the regional configuration from
\cref{sec:analysis-empirical} on a token grid of
height $h$ and width $w$.
Let $\mu\in\{0,1\}^{h\times w}$ indicate the current
editing region, with $\mu=1$ inside and $\mu=0$ outside,
and let $\tau$ denote each token's editing start time.
Attention-based region selection and expansion,
including adjustments to token start times,
are described below.

At each sampling step, we construct paired source
and target inputs.
The source input uses the forward SDE path inside
$\mu$ and fresh noisy source states outside:
\begin{equation}
Z_t^\src
=\mu\odot Z_t^{\src,\mathrm{fw}}
+(1-\mu)\odot[(1-t)\Xsrc+t\varepsilon_t],
\label{eq:structured}
\end{equation}
where $Z_t^{\src,\mathrm{fw}}$ is the forward SDE
reference from \cref{eq:forward-path},
$\varepsilon_t\sim\Normal(0,I)$ is independently
sampled at each step, and $\odot$ denotes
element-wise multiplication.
We obtain the target input by adding the current
editing displacement to the source input:
$Z_t^\tar=Z_t^\src+D_t$.

The source and target branches predict clean latents
under their respective prompts.
We use their difference $\Delta\hat x_t$ to update
the displacement $D_t$, applying the posterior rule
inside $\mu$ and the Euler rule outside:
\begin{equation}
\begin{aligned}
D_{\tnext}
={}&\mathbf{1}[t\le\tau]\odot
\Big\{\mu\odot(a_tD_t+b_t\Delta\hat x_t)\\
&+(1-\mu)\odot
\Big[\frac{\tnext}{t}D_t+
\Big(1-\frac{\tnext}{t}\Big)\Delta\hat x_t\Big]\Big\},
\end{aligned}
\label{eq:method}
\end{equation}
where $a_t,b_t$ are the posterior coefficients
from \cref{eq:posterior-coefficients}, and the
per-token indicator $\mathbf{1}[t\le\tau]$ equals
one when $t\le\tau$ and zero otherwise.
Before a token starts editing, $D_t$ remains zero,
so its source and target inputs coincide.

\paragraph{Attention-adaptive region.}
We initialize the editing region $\mu=\mu_0$
by thresholding source and target attention maps
for the subject phrase and edit-related words.

During early sampling, we expand $\mu$ to accommodate
new geometry by adding neighboring tokens whose
attention to the subject or edit-related words in
the target prompt exceeds a threshold.
Newly included tokens switch to SDE + Posterior,
with $\tau\leftarrow\max(\tau,\tnext)$ after each
step $t\to\tnext$, allowing previously inactive
tokens to begin editing at the next step.
We keep $\mu$ fixed after this early expansion phase.

\subsection{Source Anchoring}
\label{sec:method-anchor}

To further preserve subject identity and details
unrelated to the edit, we use the paired source
branch for attention-feature sharing and a single correction during sampling.

\paragraph{Shared keys and values.}
During sampling, target queries attend to the source
branch's keys and values in selected attention layers.
This gives target generation access to source features
for preserving identity and appearance.
These features come directly from the source branch
already used in the coupling.

\paragraph{Agreement-based correction.}
The expanded editing region may include content
that should remain unchanged.
We use agreement between the source and target clean
estimates as a preservation signal.
After the early expansion phase, we apply a one-time
correction:
\[
D_t\leftarrow(1-\omega)\odot D_t,
\]
where $\omega\in[0,1]^{h\times w}$ is a soft agreement
weight, set to zero outside $\mu$.
We use the typical clean-estimate difference outside
$\mu$ as a reference scale.
Within $\mu$, smaller source--target differences
relative to this scale receive larger $\omega$,
reducing the accumulated displacement more.
These tokens remain in $\mu$ and continue with
posterior updates.

\section{Experiments}
\label{sec:exp}

\phantomsection\label{sec:exp-setup}

\paragraph{Benchmarks.}
As image-editing benchmarks are dominated by appearance edits~\cite{ju2024pnpinversion}, we collected \emph{\sprite}, 491 edits that change a Shape, Pose, Rigid control, Interaction, sTate, or Expression (Supp.~\ref{supp:bench}).
We also use PIE-Bench~\cite{ju2024pnpinversion} (700 edits, mostly appearance). For video, FiVE-Bench~\cite{li2025five} (419 edits), NRVBench~\cite{qu2026nrvbench} (180 non-rigid edits), and IVEBench~\cite{chen2025ivebench} (59 subject-motion edits).

\paragraph{Metrics.}
Edit success is judged by Qwen2.5-VL~\cite{bai2025qwen25vl} on \sprite, by VQA on PIE-Bench, by action success (AS) and instruction following (IF) on IVEBench, and by GPT-5.5~\cite{openai2026gpt55} and Claude Opus 5~\cite{anthropic2026opus5} everywhere.
Id.\ and BG measure identity and background preservation; BG L1 is the background error, and Comp.\ counts successful edits that keep the identity without artifacts.
CLIP$_\mathrm{e}$ is the CLIP~\cite{radford2021clip} score of the edited region.  $S_\mathrm{instr}$ is the instruction score (Supp.~\ref{supp:metrics}).

\paragraph{Baselines.}
On images, we compare with the inversion-free FlowEdit~\cite{kulikov2025flowedit} and Sync-SDE~\cite{zhang2026syncsde}, the inversion-based RF-Solver-Edit~\cite{wang2025rfsolver} and FireFlow~\cite{deng2025fireflow}, and mask-based SAM-Flow~\cite{cui2026samflow}, all on FLUX.1-dev~\cite{bfl2024flux} with their released code and settings. On video, we compare with Wan-Edit~\cite{li2025five} on Wan2.1 and editors reported in FiVE-Bench and NRVBench. \method's hyperparameters were set on images from previous papers~\cite{kulikov2025flowedit, chen2026synps}, none of which appear in \sprite.
SynPS~\cite{chen2026synps} cannot edit a given image: it starts from the initial noise of a source it generates.
DynaEdit~\cite{kulikov2026dynaedit} sets its hyperparameters for each video and has not released them. We do not compare against them.

\phantomsection\label{sec:exp-image}\label{sec:exp-video}

\begin{figure*}[t]
{\ueinnerfloats\input{fig/qual_video}}
\par\vspace{12pt}
{\ueinnerfloats\input{fig/oneword}}
\end{figure*}

\paragraph{Qualitative evaluation.}
\Cref{fig:teaser} shows \method on images, videos, sound, speech, and 3D assets, from pose and shape changes to adding, removing, moving, and swapping objects and rewriting scene text, while preserving the source. 
In \cref{fig:qual}, \method is the only method that makes all five image edits and preserves the rest. For example, it stands the panda on its hind legs and moves the man's hands onto his hips, whereas every baseline keeps the source pose. In \cref{fig:oneword}, it rewrites a sign's word in the sign's typeface and keeps the facade, which FlowEdit, RF-Solver-Edit, and FireFlow redraw.
On video, \method makes the dancer spin in a pirouette, turns the fox into a robot, and turns the woman into a cat, while Wan-Edit fails all three edits.
More comparisons are in Supp.~\ref{supp:teaser-cmp},~\ref{supp:qual-nre}, and~\ref{supp:video-more}.

\begin{table}[t]
{\ueinnerfloats\begin{table}[t]
\centering
\caption{\textbf{Image editing comparison} with prior methods on \sprite and PIE-Bench. Bold: best; \underline{underline}: second-best.}
\label{tab:main}
\definecolor{qwenbase}{HTML}{615CED}   %
\definecolor{gptbase}{HTML}{2A6BD8}    %
\definecolor{claudebase}{HTML}{D97757} %
\colorlet{qwenfg}{qwenbase!78!black}
\colorlet{gptfg}{gptbase!80!black}
\colorlet{claudefg}{claudebase!75!black}
\providecommand{\jq}[1]{\textcolor{qwenfg}{#1}}
\providecommand{\jg}[1]{\textcolor{gptfg}{#1}}
\providecommand{\jl}[1]{\textcolor{claudefg}{#1}}
\setlength{\aboverulesep}{0pt}\setlength{\belowrulesep}{0pt}\setlength{\extrarowheight}{1.2pt}
\uetabstyle%
\begin{tabularx}{\linewidth}{l *{4}{Y} c *{6}{Y}}%
\toprule
 & & \multicolumn{6}{c}{\sprite} & \multicolumn{4}{c}{PIE-Bench} \\
\cmidrule(lr){3-8}\cmidrule(l){9-12}
 & NFE & \jq{Qwen\un{\%}\,$\uparrow$} & \jg{GPT\un{\%}\,$\uparrow$} & \jl{Opus\un{\%}\,$\uparrow$} & Comp.\un{\%}\,$\uparrow$ & Id.\un{\%}\,$\uparrow$ & BG\un{dB}\,$\uparrow$ & \jq{VQA\un{\%}\,$\uparrow$} & \jg{GPT\un{\%}\,$\uparrow$} & \jl{Opus\un{\%}\,$\uparrow$} & BG\un{dB}\,$\uparrow$ \\
\midrule
RF-Solver-Edit & 60 & 48.7 & 44.0 & 43.8 & 37.1 & 39.9 & 11.59 & \underline{64.1} & \underline{54.9} & \underline{61.3} & \underline{12.65} \\
FireFlow & 18 & 43.8 & 38.1 & 38.7 & 32.5 & 35.8 & 10.45 & 63.4 & 52.0 & 58.6 & 12.18 \\
FlowEdit & 48 & \underline{50.1} & \underline{44.2} & \underline{45.0} & \underline{41.2} & \underline{42.2} & 11.13 & 59.4 & 46.6 & 52.0 & 11.09 \\
Sync-SDE & 48 & 40.5 & 31.2 & 33.6 & 21.3 & 32.7 & 11.23 & 52.7 & 38.4 & 45.9 & 10.81 \\
SAM-Flow & 852 & 38.9 & 30.3 & 33.4 & 27.8 & 32.7 & \underline{12.69} & 44.5 & 21.0 & 26.6 & 8.86 \\
\midrule
\ours \textbf{\method (Ours)} & 38 & \textbf{78.0} & \textbf{69.0} & \textbf{73.7} & \textbf{52.7} & \textbf{61.8} & \textbf{18.19} & \textbf{70.0} & \textbf{65.6} & \textbf{72.9} & \textbf{14.72} \\
\bottomrule
\end{tabularx}
\end{table}
}
\par\vspace{10pt}
{\ueinnerfloats\begin{table}[t]
\centering
\caption{\textbf{Video editing comparison} with Wan-Edit on FiVE-Bench, NRVBench, and IVEBench motion.}
\label{tab:video}
\definecolor{gptbase}{HTML}{10A37F}    %
\definecolor{claudebase}{HTML}{D97757} %
\colorlet{gptfg}{gptbase!72!black}
\colorlet{claudefg}{claudebase!75!black}
\definecolor{gptbluebase}{HTML}{2A6BD8}  %
\colorlet{gptnavy}{gptbluebase!80!black}
\providecommand{\jg}[1]{\textcolor{gptnavy}{#1}}
\providecommand{\jl}[1]{\textcolor{claudefg}{#1}}
\definecolor{qwenbase}{HTML}{615CED}   %
\colorlet{qwenfg}{qwenbase!78!black}
\providecommand{\jb}[1]{\textcolor{qwenfg}{#1}}
\providecommand{\vidgray}[1]{\textcolor{black!55}{#1}}
\uetabstyle%
\begin{tabularx}{\linewidth}{l *{6}{Y}}
\toprule
\textbf{FiVE-Bench}, 419 edits & NFE & \jb{FiVE-Acc\un{\%}\,$\uparrow$} & \jg{GPT-5.5\un{\%}\,$\uparrow$} & \jl{Opus 5\un{\%}\,$\uparrow$} & CLIP$_\mathrm{e}$\un{$\times$100}\,$\uparrow$ & BG\un{dB}\,$\uparrow$ \\
\midrule
Wan-Edit & 140 & 46.97 & 42.00 & 54.65 & 21.23 & 10.77 \\
\ours \textbf{\method (Ours)} & 76 & \textbf{53.05} & \textbf{56.32} & \textbf{67.78} & \textbf{21.60} & \textbf{14.43} \\
\midrule
\textbf{NRVBench}, 180 edits & NFE & \jb{NRVE-Acc\un{\%}\,$\uparrow$} & \jg{GPT-5.5\un{\%}\,$\uparrow$} & \jl{Opus 5\un{\%}\,$\uparrow$} & $S_\mathrm{instr}$\un{0--100}\,$\uparrow$ & BG\un{dB}\,$\uparrow$ \\
\midrule
Wan-Edit & 140 & 34.55 & 11.67 & 21.67 & 65.18 & 3.77 \\
\ours \textbf{\method (Ours)} & 76 & \textbf{34.63} & \textbf{41.11} & \textbf{48.89} & \textbf{73.43} & \textbf{10.62} \\
\midrule
\textbf{IVEBench motion}, 59 edits & NFE & \jb{AS\un{\%}\,$\uparrow$} & \jg{GPT-5.5\un{\%}\,$\uparrow$} & \jl{Opus 5\un{\%}\,$\uparrow$} & IF\un{1--4}\,$\uparrow$ & BG\un{dB}\,$\uparrow$ \\
\midrule
Wan-Edit & 140 & 4.5 & 7.9 & 9.0 & 1.61 & 2.33 \\
\ours \textbf{\method (Ours)} & 76 & \textbf{34.5} & \textbf{35.6} & \textbf{54.8} & \textbf{2.31} & \textbf{7.87} \\
\bottomrule
\end{tabularx}
\end{table}}
\end{table}

\paragraph{Quantitative evaluation.}
\method is the best method on every metric of \cref{tab:main}, on both benchmarks and under all three judges.
On \sprite, it succeeds on 78.0\% of the edits, 27.9 points above the best baseline, FlowEdit.
It also preserves the source best: \method has the highest Comp., Id., and BG.
On shape and state edits, which need a new layout, ours succeeds about four times as often as FlowEdit (Supp.~\cref{tab:supp-cat}).
On PIE-Bench, mostly appearance edits, \method still leads with 70.0 VQA against 64.1 for RF-Solver-Edit.
It also takes less time per edit than every image baseline except FireFlow, which runs 18 NFE (Supp.~\cref{tab:supp-cost}).
On video, \method scores above Wan-Edit on every metric (\cref{tab:video}); its FiVE-Bench accuracy of 53.05 is above every editor the benchmark reports (Supp.~\cref{tab:supp-video-published}).
On NRVBench, GPT-5.5 judges 41.1\% of its edits successful vs. 11.7\% for Wan-Edit; on IVEBench motion edits, AS rises from 4.5\% to 34.5\%.

\paragraph{User study.}
In a blinded study, 15 participants preferred \method to a baseline in 76\% of 915 image and 93\% of 510 video comparisons ($p<0.001$; Supp.~\ref{supp:userstudy}).

\phantomsection\label{sec:apps}

\paragraph{Generalization to other models, modalities, and tasks.}
The update needs only a velocity field, information from two prompts, and a noise path, so we also apply it unchanged to other generators (\cref{fig:teaser}, Supp.~\ref{supp:apps}).
On SD3~\cite{esser2024sd3}, it passes 70.7\% of a 41-edit \sprite subset against 56.1\% for FlowEdit at its released SD3 settings, with 72 NFE against 132.
On the audio and speech generators TangoFlux~\cite{hung2024tangoflux} and F5-TTS~\cite{chen2025f5tts}, we test whether the attention region matters by comparing with the same code run with an empty region ($\mu\equiv0$): tempo and rhythm edits pass on 10/14 runs against 2/14, and word edits on 43/44 against 19/44.
These are demonstrations of the update on other modalities, not comparisons with dedicated audio editors.
On TRELLIS~\cite{xiang2025trellis}, it changes the appearance and parts of generated 3D assets (Supp.~\cref{fig:supp-3d-more}).
On images, it also handles sequential editing, moving a subject to a given point, and several subjects or edits at once (Supp.~\cref{fig:supp-apps-a,fig:supp-apps-b}).

\phantomsection\label{sec:exp-design}

\begin{table}[t]
{\ueinnerfloats\begin{table}[t]
\centering
\caption{\textbf{Region and noise.}
In the region, the forward path with the posterior step edits most. Entries: region\,/\,rest; indep.: the target's own noise, not shared with the source. Evaluated on \sprite.}
\label{tab:design}
\uetabstyle%
\begin{tabularx}{\linewidth}{l *{6}{Y}}
\toprule
 & noise & step & Edit\un{\%}\,$\uparrow$ & Id.\un{\%}\,$\uparrow$ & BG\un{dB}\,$\uparrow$ & BG L1\un{0--255}\,$\downarrow$ \\
\midrule
\multicolumn{7}{@{}l}{\textbf{Inside the region}} \\
\quad fresh noise & fresh\,/\,fresh & post.\,/\,ODE & 73.3 & 58.4 & \underline{18.36} & 10.78 \\
\quad fresh noise, ODE step & fresh\,/\,fresh & ODE\,/\,ODE & 63.7 & 51.4 & 17.01 & \textbf{8.38} \\
\quad independent & indep.\,/\,fresh & post.\,/\,ODE & 72.5 & 54.1 & 14.35 & 21.09 \\
\quad ODE step & forward\,/\,fresh & ODE\,/\,ODE & 73.7 & 58.7 & \textbf{18.83} & \underline{9.72} \\
\midrule
\multicolumn{7}{@{}l}{\textbf{Outside the region}} \\
\quad forward path, posterior step & forward\,/\,forward & post.\,/\,post. & \underline{76.6} & \underline{60.0} & 16.40 & 16.37 \\
\midrule
\ours \textbf{\method (Ours)} & forward\,/\,fresh & post.\,/\,ODE & \textbf{78.0} & \textbf{61.8} & 18.19 & 12.78 \\
\bottomrule
\end{tabularx}
\end{table}
}
\par\vspace{10pt}
{\ueinnerfloats\begin{table}[t]
\centering
\caption{\textbf{Source anchoring and region expansion.}
Removing either part of source anchoring does not significantly change edit success and loses preservation; without expansion of $\mu$, edit success drops, and its lower BG L1 comes from editing less.}
\label{tab:anchor}
\uetabstyle%
\begin{tabularx}{\linewidth}{l *{4}{Y}}
\toprule
 & Edit\un{\%}\,$\uparrow$ & Id.\un{\%}\,$\uparrow$ & BG\un{dB}\,$\uparrow$ & BG L1\un{0--255}\,$\downarrow$ \\
\midrule
no shared $K$ and $V$ & \textbf{78.2} & 58.5 & 17.48 & 14.78 \\
no scaling back at $t_r$ & 76.8 & \underline{60.7} & 17.18 & 15.01 \\
no expansion of $\mu$ & 73.7 & 58.1 & \underline{17.93} & \textbf{11.47} \\
\midrule
\ours \textbf{\method (Ours)} & \underline{78.0} & \textbf{61.8} & \textbf{18.19} & \underline{12.78} \\
\bottomrule
\end{tabularx}
\end{table}
}
\end{table}

\paragraph{Ablation: coupling.}
Each choice in the coupling matters. As shown in \cref{tab:design}, for \emph{Inside the region}, fresh noise instead of the forward path lowers edit success and Id.\ by 4.7 and 3.4 points, the ODE step instead of the posterior step by 4.3 and 3.1 points, and both replacements together by 14.3 and 10.4 points.
Independent target noise instead of the noise shared with the source copy lowers edit success by 5.5 points, Id.\ by 7.7 points, and BG by 3.8\,dB.
\emph{Outside the region}, the forward path with the posterior step instead of fresh noise with the ODE step lower edit success by 1.4 points, Id. by 1.8 points, and BG by 1.8 dB.

\paragraph{Ablation: region and probe time.}
Among the settings we tested, \method's split start works best: the region edits most when it starts at pure noise and expands, the rest preserves more when it starts later, and the region is robust to its probe time.
In \cref{tab:anchor}, without expansion, $\mu$ stays at $\mu_0$ and edit success falls from 78.0\% to 73.7\%, most on interactions ($-11.7$) and human poses ($-9.0$), where a raised arm or a held object reaches beyond $\mu_0$; BG does not change significantly.
\method is robust to the probe time as shown in Supp.~\cref{tab:probe}: replacing the default probes at $t=0.7$ and $0.5$ by a single probe time anywhere from $t=0.9$ to $0.3$ (one source and one target call), edit success stays within 3 points and BG within 0.6\,dB of the default.

\paragraph{Ablation: source anchoring.}
Both parts of source anchoring improve preservation at no significant cost in edit success, and they work because they build on \method's coupling.
In \cref{tab:anchor}, without the scaling-back step, BG is 1.01\,dB lower; without shared keys and values, Id.\ is 3.3 points lower.
\method is also robust to the scaling-back time (see Supp.~\cref{tab:probe}): for $t_r\in[0.6,0.8]$, edit success stays within 1 point and BG within 0.21\,dB of the default.

\section{Analysis}
\label{sec:exp-analysis}
\begin{figure*}[t]
\centering
\includegraphics[width=\textwidth]{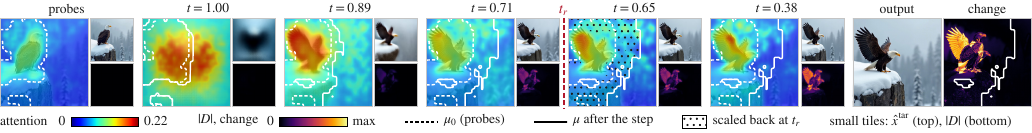}
\caption{\textbf{The edit stays where attention points.}
Prompt: ``wings folded''$\to$``wings open and spread''.}
\label{fig:attn}
\end{figure*}

\paragraph{Attention map.}
Without an external segmenter, \method's attention locates the edit accurately. In \cref{fig:attn}, before any editing step, at $\mu_0$, attention on the edited words already covers the eagle (35\% of the whole image), then expands to 66\% to make room for the spread wings. At $t_r$, $\mu$ is fixed; where the two copies agree, in the sky and on the rock, tokens stay in $\mu$ but move most of the way back toward the source ($w>0.5$). The rest of $\mu$ ($w\le0.5$) covers 29\% of the image yet holds 97.5\% of the change.

\begin{figure*}[t]
\centering
\includegraphics[width=0.85\textwidth]{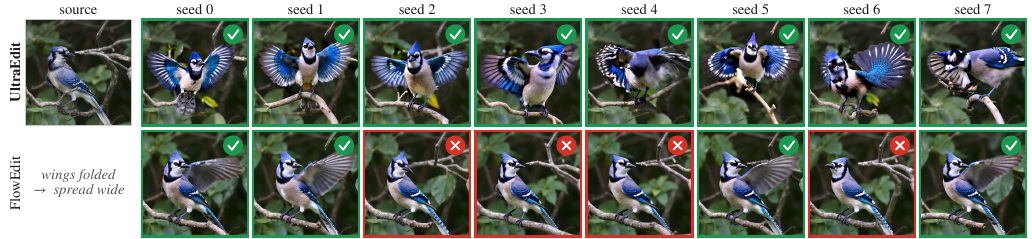}
\caption{\textbf{Seed diversity.}
With \method, the seed chooses the pose.
\method spreads the wings in all eight seeds, each time differently; FlowEdit succeeds in four, raising the same wing the same way.
Frame and badge: judged edit success.
Settings: ``Make the blue jay spread its wings wide''; FlowEdit as released.}
\label{fig:diversity}
\end{figure*}

\begin{figure*}[t]
\centering
\includegraphics[width=\textwidth,trim=0 93.451bp 0 0,clip]{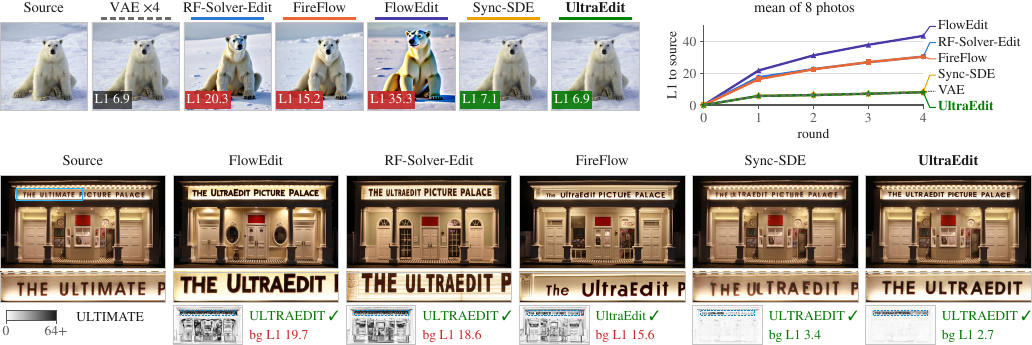}
\caption{\textbf{Zero edits in a row.}
\method keeps what the prompt does not change.
Four zero edits in a row (round $4$ shown, L1 to the source); the plot averages eight real photos.
Green: a kept source (round-$4$ L1 within $3$ of the VAE round trips); red: a changed source.
Settings: target prompt $=$ source prompt; baselines at their released settings.}
\label{fig:preserve}
\end{figure*}

\paragraph{Seed diversity.}
With \method, the seed chooses how a structural edit is carried out, while the background stays as close to the source as with FlowEdit.
In \cref{fig:diversity}, \method spreads the blue jay's wings in all eight seeds, each time in a different pose; FlowEdit succeeds in four, with the same pose.
Background L1 to the source is 10.4 vs.\ 12.9 for FlowEdit. Supp.~\ref{supp:seeds} reports seven more edits.

\paragraph{Zero edits.}
After 4 consecutive zero edits (target prompt = source prompt), \method stays close to the sources, with a mean L1 distance of 8.0 gray levels, the same as four VAE round trips alone, whereas FireFlow, which inverts the source and regenerates it, deviates by 30.4 gray levels (\cref{fig:preserve}).

\section{Conclusion and limitations}
\label{sec:concl}

We introduced \method, an inversion-free and training-free method for editing with pretrained flow models.
It chooses the start time and the reference-path coupling per token from the model's attention: the attention region starts from pure noise on a forward path, and the rest starts later from fresh noise.
Our analysis shows that an early start enables structural edits, and that choosing per token preserves more of the rest.
On \sprite, \method succeeds on 78.0\% of the edits versus 50.1\% for FlowEdit, and it applies to video, sound, speech, and 3D generators.
While relying on the pretrained generator avoids training, it can become a limitation when the edited image must be reasoned out from the prompt, as discussed in Supp.~\cref{sec:limitation}.
{\small\bibliographystyle{ieeenat_fullname}\bibliography{main}}
\clearpage
\providecommand{\Vd}{\mathbf{V}}
\providecommand{\qE}{q_{E}}
\clearpage
\ifdefined\suppstart\suppstart\fi
\appendix
\newcounter{supppart}
\newcommand{\suppparthead}[1]{\refstepcounter{supppart}\pdfbookmark[0]{Part \Roman{supppart}: #1}{supppart.\arabic{supppart}}%
  {\centering{\small\ueeyebrow{Part \Roman{supppart}}\hspace{0.6em}\raisebox{0.3ex}{\color{ueAccent}\rule{1.6em}{0.5pt}}}\par\vspace{4pt}%
   {\sffamily\interextrabold\fontsize{15}{19}\selectfont\color{ueInk}#1\par}%
   \vspace{6pt}\uegradrule{0.9in}{1.1pt}\par}\vspace{1.1em}}%
\newcommand{\supppart}[1]{\clearpage\suppparthead{#1}}
\newcommand{\supptocpart}[1]{\par\vspace{7pt}\noindent{\sffamily\intersemibold\color{ppInk}#1}\par\vspace{2pt}}
\newcommand{\supptocitem}[2]{\noindent\hspace*{1.5em}\makebox[1.6em][l]{\textbf{\ref{#1}}}#2\dotfill\pageref{#1}\par\vspace{1pt}}
\onecolumn
{%
  \centering
  \begin{tcolorbox}[enhanced, width=0.78\textwidth, arc=10pt, boxrule=0pt, toprule=1.8pt,
    left=0.25in, right=0.25in, top=0.16in, bottom=0.18in,
    frame code={\shade[shading=ue@face, rounded corners=10pt] (frame.south west) rectangle (frame.north east);},
    interior code={\fill[ueTint, rounded corners=9pt] (interior.south west) rectangle (interior.north east);}]%
  {\centering{\small\ueeyebrow{Supplementary Material}\hspace{0.6em}\raisebox{0.3ex}{\color{ueAccent}\rule{1.6em}{0.5pt}}\par}\vspace{0.5em}
   {\fontsize{24}{26}\selectfont\uewordmark{supp}{\sffamily\interheavy}\par}\vspace{0.2em}
   {\sffamily\bfseries\fontsize{12}{15}\selectfont\color{ppInk}Text\uehyph Guided Editing with Adaptive Coupling\par}}\vspace{0.7em}
  {\color{ppAccent!45}\hrule height 0.4pt}\vspace{2pt}
  \small
  \supptocpart{Part I: Teaser comparison}
  \supptocitem{supp:teaser-cmp}{Teaser comparison with baseline editors}
  \supptocpart{Part II: Background and theory}
  \supptocitem{supp:proofs}{Proofs for \cref{sec:analysis}}
  \supptocitem{supp:when}{Proof of \cref{prop:reach}}
  \supptocpart{Part III: Image experiments: Setup and quantitative results}
  \supptocitem{supp:impl}{Implementation details}
  \supptocitem{supp:bench}{The \sprite benchmark}
  \supptocitem{supp:metrics}{Metrics}
  \supptocitem{supp:design}{Ablations}
  \supptocitem{supp:more-baselines}{Inversion-based editors}
  \supptocitem{supp:cost}{Cost per edit}
  \supptocitem{supp:userstudy}{User study}
  \supptocpart{Part IV: Image qualitative results, limitation, and failure cases}
  \supptocitem{supp:qual-nre}{More image comparisons}
  \supptocitem{supp:zero}{Zero edits}
  \supptocitem{sec:limitation}{Limitations and failure cases}
  \supptocpart{Part V: Video experiments}
  \supptocitem{supp:video}{Video editor and benchmarks}
  \supptocitem{supp:video-more}{More video edits}
  \supptocpart{Part VI: Other generators and tasks}
  \supptocitem{supp:apps}{Other generators and tasks}
  \end{tcolorbox}\par\vspace{1.8em}
  \suppparthead{Teaser comparison}}

\section{Teaser comparison with baseline editors}
\label{supp:teaser-cmp}

\begin{figure*}[tp]
\centering
\includegraphics[width=\textwidth]{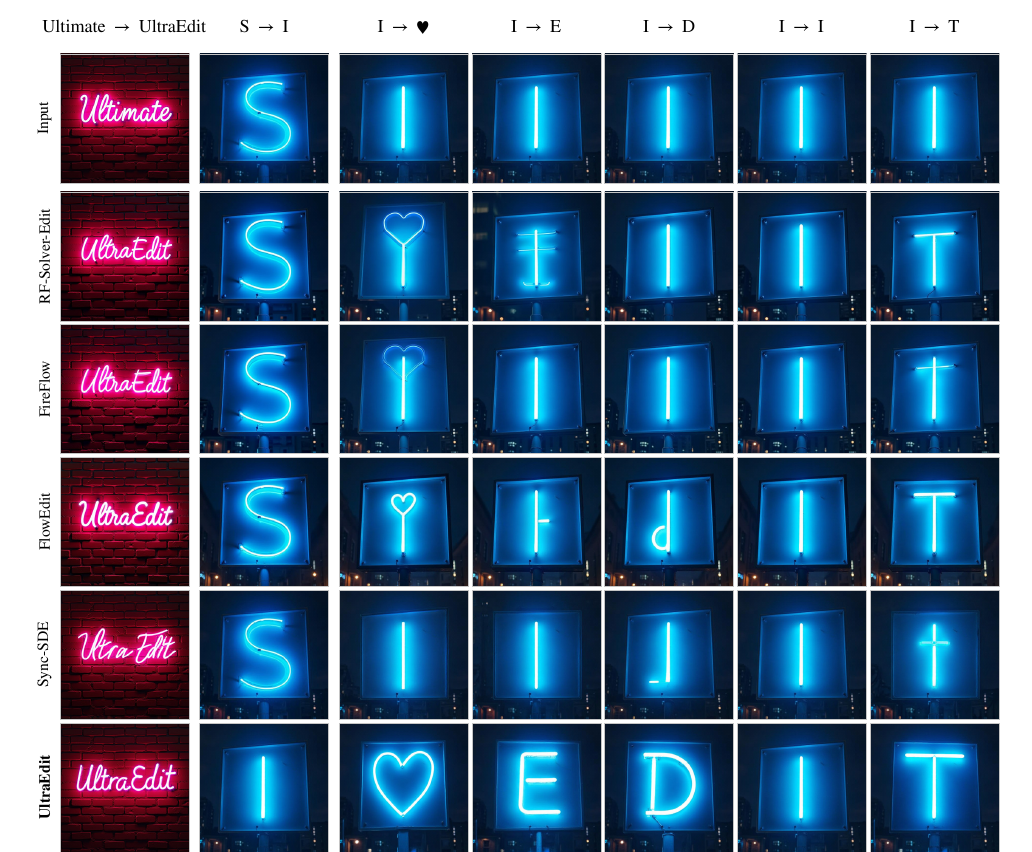}
\caption{\textbf{Teaser comparisons: neon sign with a word or a letter.}
No other editor turns the S into an I.
On the I, Sync-SDE keeps the I for the heart and the E, adds only a short foot for the D, and draws a cross for the T; FireFlow keeps the I apart from a faint heart on a stem and a faint cross; RF-Solver-Edit puts a heart on top of the I, crosses it with two short bars for the E, and keeps the I for the D; FlowEdit draws a small heart on a stem, a single short bar for the E, and a lowercase d.
On the sign, Sync-SDE splits the word into ``Ultra Edit''.
\method replaces the tube with the full target shape in every column and keeps the I when the target is the I itself (I $\rightarrow$ I).
\textbf{Input} row: a generated neon sign, the neon S of SynPS, and the I that \method made from the S.}
\label{fig:supp-tcmp-neon}
\end{figure*}

\begin{figure*}[tp]
\centering
\includegraphics[width=\textwidth]{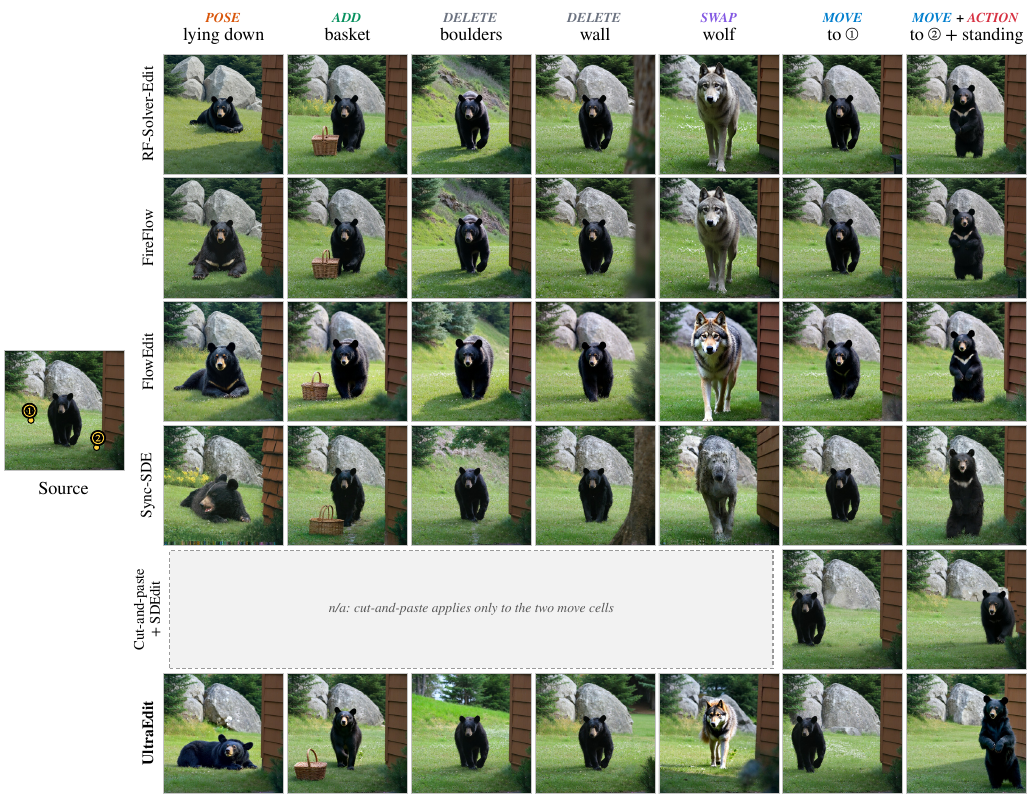}
\caption{\textbf{Teaser comparisons: bear.}
No prompt-only editor moves the bear, and for the second point they stand it up where it is; cut-and-paste moves it but cannot make it stand.
\method moves the bear to the first point and, for the second, moves it toward the point and stands it up.
In the other columns, FireFlow seats the bear instead of laying it down, RF-Solver-Edit, FireFlow, and Sync-SDE put a tree trunk where the wall was, and RF-Solver-Edit, FireFlow, and FlowEdit draw a wolf much larger than the bear; \method lays the bear down, opens the lawn where the wall stood, and draws a wolf of the bear's size.
For the moves, the prompt-only editors get the target position in the prompt; cut-and-paste uses FLUX SDEdit inpainting of the hole (\cref{supp:apps}, moving to a point).}
\label{fig:supp-tcmp-bear}
\end{figure*}

\begin{figure*}[tp]
\centering
\includegraphics[width=\textwidth]{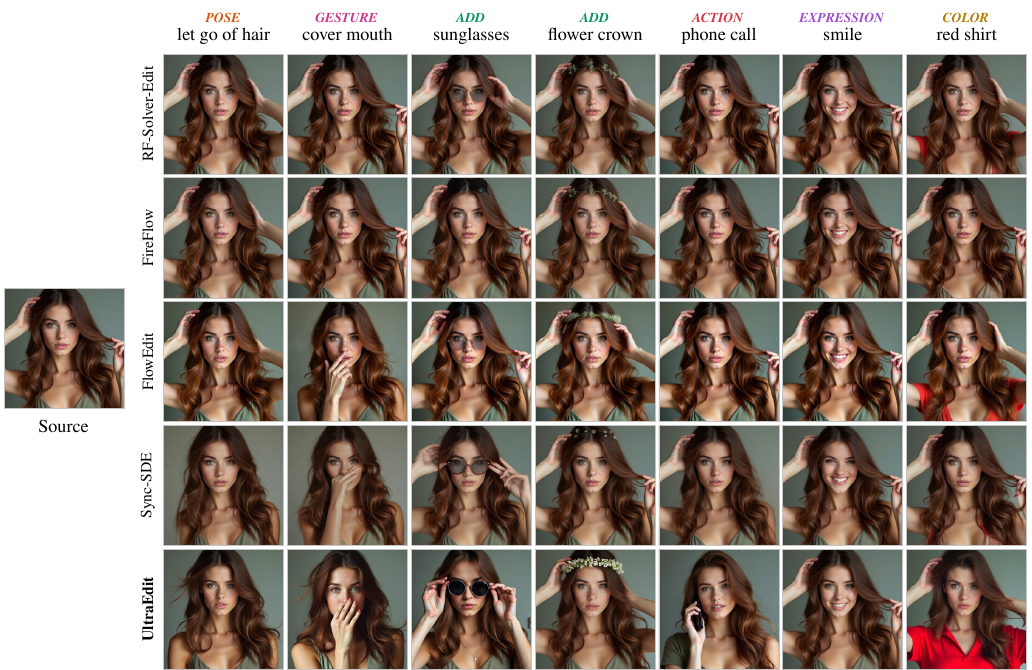}
\caption{\textbf{Teaser comparisons: woman.}
No other editor makes the phone call; RF-Solver-Edit and FireFlow do not cover the mouth, and FireFlow puts the sunglasses on top of her head.
\method makes all seven edits.}
\label{fig:supp-tcmp-woman}
\end{figure*}

\begin{figure*}[tp]
\centering
\includegraphics[width=\textwidth]{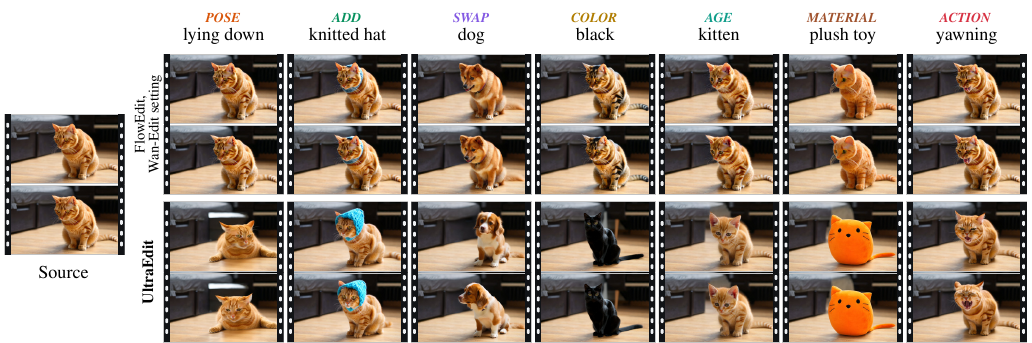}
\caption{\textbf{Teaser comparisons: video of a tabby cat.}
In Wan-Edit's setting the cat neither lies down nor becomes a kitten, the hat becomes a scarf, the color turns black only in patches, and the plush toy is still a cat.
\method makes all seven edits.
\textbf{Top}: FlowEdit in Wan-Edit's released setting ($50$ steps, shift $5$, guidance $5/10$) in our Wan code, as this generated clip has no released Wan-Edit output.}
\label{fig:supp-tcmp-cat}
\end{figure*}

\begin{figure*}[tp]
\centering
\includegraphics[width=\textwidth]{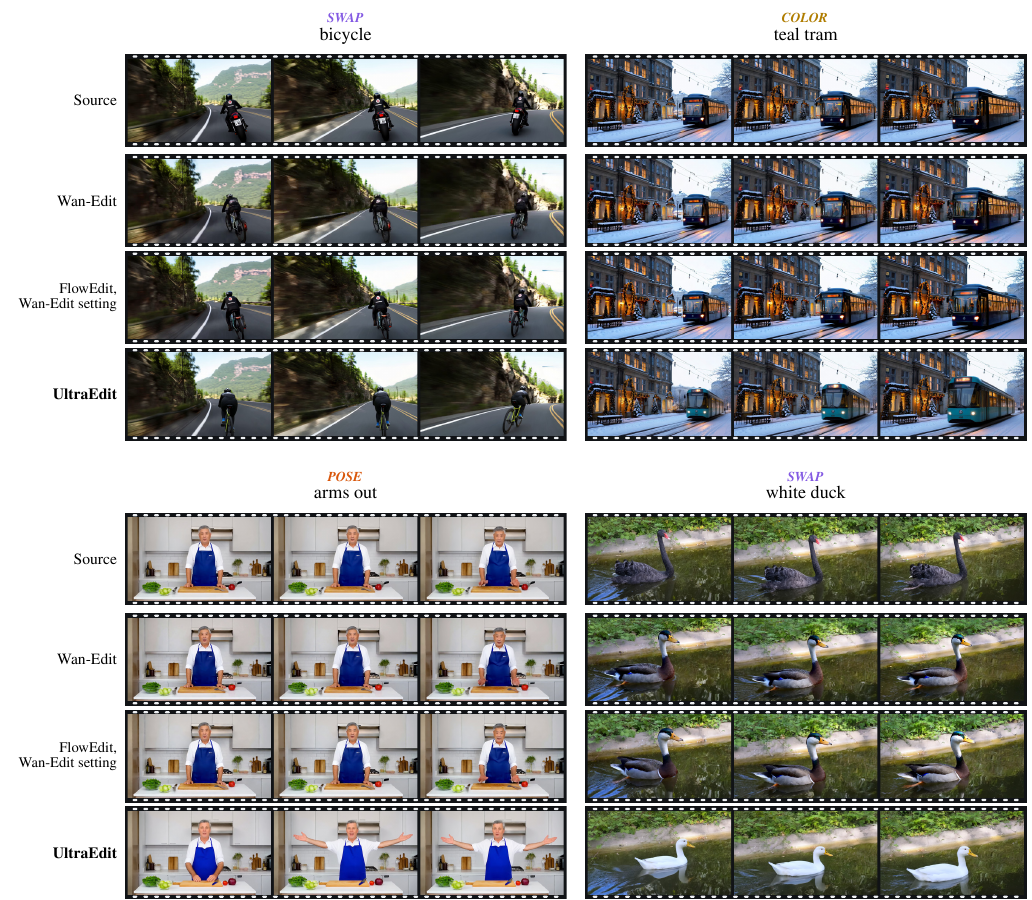}
\caption{\textbf{Teaser comparisons: video benchmark clips (1 of 2).}
Neither Wan-Edit nor FlowEdit in its setting makes the tram teal or stretches the arms out, and both turn the swan into a mallard instead of a white duck.
\method makes all four edits.
Each strip shows the first, the middle (shown in the teaser), and the last frame.
Wan-Edit: FiVE-Bench's released output on the FiVE-Bench clips, its released code on the others.}
\label{fig:supp-tcmp-veval}
\end{figure*}

\begin{figure*}[tp]
\centering
\includegraphics[width=\textwidth]{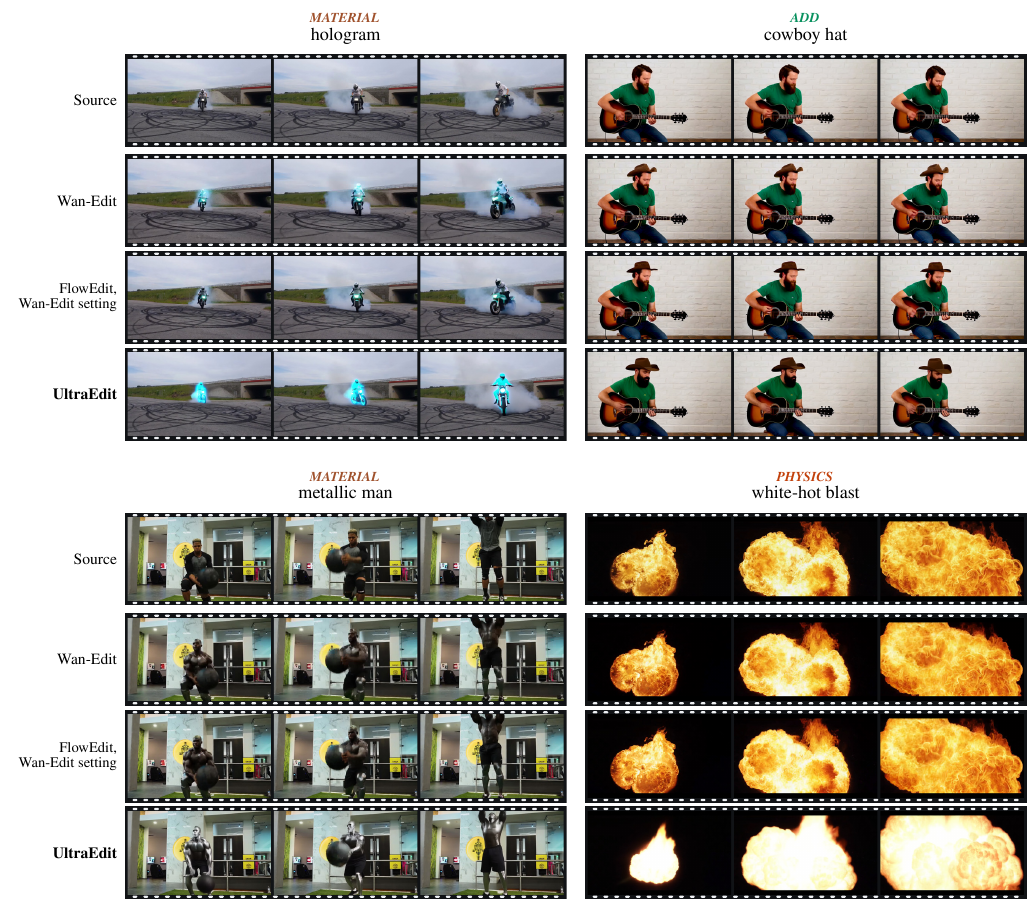}
\caption{\textbf{Teaser comparisons: video benchmark clips (2 of 2).}
Neither Wan-Edit nor FlowEdit in its setting turns the blast white-hot, and both make the hologram and the metallic man only in part.
\method makes all four edits.
Frames and Wan-Edit outputs as in \cref{fig:supp-tcmp-veval}.}
\label{fig:supp-tcmp-veval-b}
\end{figure*}

\begin{figure*}[tp]
\centering
\includegraphics[width=\textwidth]{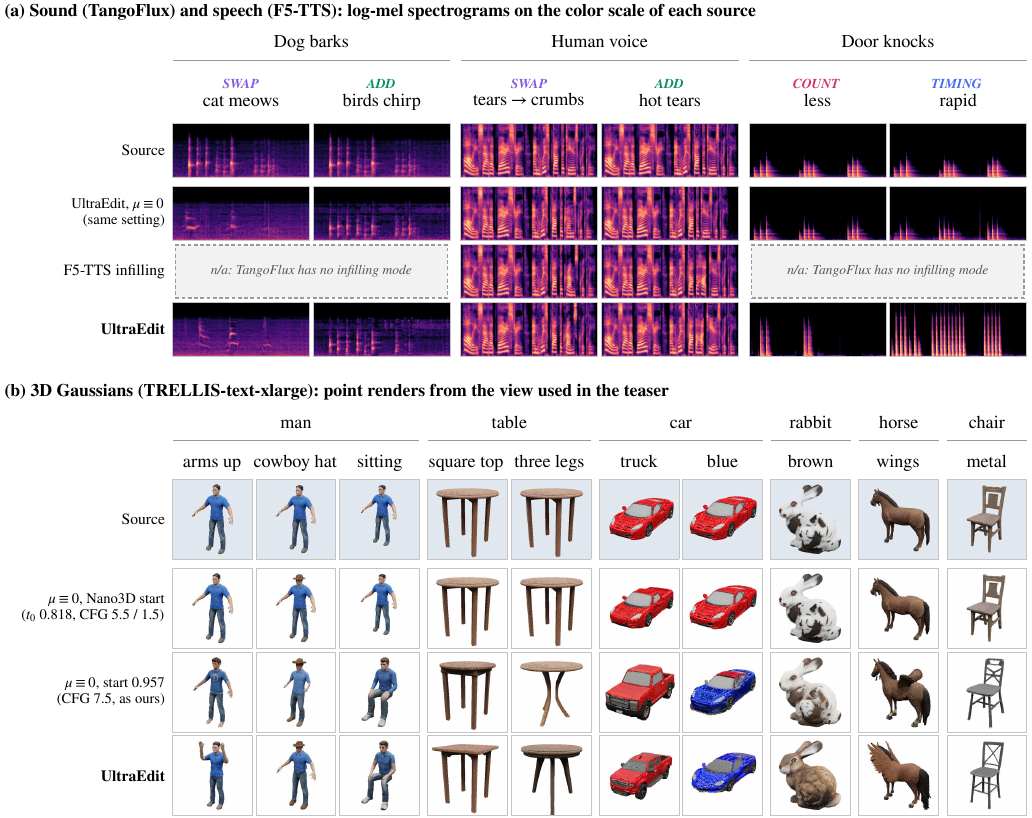}
\caption{\textbf{Teaser comparisons: sound, speech, and 3D.} $\mu\equiv0$ is the same code with an empty attention region. 
\textbf{(a)}~$\mu\equiv0$ does not add the birds, fails ``hot tears'' on speech, and keeps the count and timing of the knocks.
\method removes a group of knocks and plays them faster, each knock still sharp.
\textbf{(b)}~From Nano3D's start ($0.818$), $\mu\equiv0$ adds only the hat; from $0.957$ it makes more edits but does not raise the arms.
Only \method raises the arms, makes the square top and the spread wings, and turns the whole rabbit brown; $\mu\equiv0$ leaves the rabbit largely white; at most, a few patches change to brown.}
\label{fig:supp-tcmp-audio3d}
\end{figure*}

For a qualitative comparison with existing methods, we reproduce each example
of \cref{fig:teaser} using competing editors in
\cref{fig:supp-tcmp-neon,fig:supp-tcmp-bear,fig:supp-tcmp-woman,fig:supp-tcmp-cat,fig:supp-tcmp-veval,fig:supp-tcmp-veval-b,fig:supp-tcmp-audio3d}.
Each figure adopts the same inputs, prompts, column layout, and labels as the
teaser, and the last row shows the corresponding result of our method.
All image editing baselines are evaluated with their officially released
configurations on FLUX.1-dev, consistent with the protocol of \cref{tab:main}.
Quantitative results are reported in \cref{tab:main,tab:video} and
\cref{supp:apps}.

\supppart{Background and theory}

\section{Proofs for \cref{sec:analysis}}
\label{supp:proofs}

\paragraph{Notation.}
We use the source and target inputs, displacement $D_t$, and affine step of \cref{sec:prelim,sec:analysis}.
For a prompt $c$, $p_c$ is the clean-latent law, $p_{c,t}$ is the law of $z_t=(1-t)x+t\varepsilon$ for independent $x\sim p_c$ and $\varepsilon\sim\Normal(0,I)$, and $p_c(x\mid z_t=z)$ is its clean posterior.
Write $\psrc=p_{\csrc}$ and $\ptar=p_{\ctar}$.
An exact clean estimate satisfies $\hat x(z,t,c)=\E_{p_c}[x\mid z_t=z]$.
For $h=t-\tnext$, let $\hat x_t^\src=\hat x(Z_t^\src,t,\csrc)$, $\hat x_t^\tar=\hat x(Z_t^\tar,t,\ctar)$, and $\Delta\hat x_t=\hat x_t^\tar-\hat x_t^\src$.
Writing $\hat\varepsilon=z+(1-t)v$ gives
\begin{equation}
z=(1-t)\hat x+t\hat\varepsilon,\qquad v=\hat\varepsilon-\hat x.
\label{eq:supp-split}
\end{equation}
The Euler step has $(a_t,b_t)=(\tnext/t,1-\tnext/t)$; the posterior coefficients are given in \cref{eq:posterior-coefficients}.
In both cases, $\Phi_t^c(z)=a_tz+b_t\hat x(z,t,c)$ and $\Xtar=\Xsrc+D_0$.
For probability laws $P,Q$, we use $\|P-Q\|_{\mathrm{TV}}=\mathrm{TV}(P,Q)=\sup_A|P(A)-Q(A)|$, equal to $\tfrac12\sum_j|P_j-Q_j|$ for discrete laws.

\subsection{Target dynamics}
\label{supp:proofs-target}

\begin{proof}[Proof of \cref{lem:target}]
For any reference process, substitute $Z_t^\tar=Z_t^\src+D_t$ and $r_t=Z_{\tnext}^\src-\Phi_t^\src(Z_t^\src)$ to obtain
\[
Z_{\tnext}^\tar-\Phi_t^\tar(Z_t^\tar)-r_t
=D_{\tnext}-a_tD_t-b_t(\hat x_t^\tar-\hat x_t^\src).
\]
This vanishes if and only if \cref{eq:offset} holds, for either coefficient pair.
\end{proof}

\addtocounter{proposition}{1}

\subsection{The forward path and target sampling dynamics}
\label{supp:proofs-sync}

We use the forward SDE of Rout et al.~\cite[Lemma~A.2]{rout2025rfinversion}, which has the same marginals as rectified flow, together with the synchronous coupling used in Sync-SDE~\cite[Sec.~5]{zhang2026syncsde}.
The construction builds on diffusion time reversal~\cite{anderson1982reverse} and related shared-noise constructions~\cite{wu2023cyclediffusion,nie2024blessing}.
We derive the posterior discretization and start-time
convergence guarantee, and adapt the Brownian-coupling
argument of Sync-SDE~\cite[Sec.~6]{zhang2026syncsde}
to partial noise sharing.

\paragraph{Forward process and posterior coefficients.}
For $0\le t<1$, set $g(t)^2=2t/(1-t)$ and
\begin{equation}
dZ_t^\src=-\frac{Z_t^\src}{1-t}\,dt+g(t)\,dW_t,\qquad Z_0^\src=\Xsrc,
\label{eq:supp-forward}
\end{equation}
where $W$ is Brownian and independent of $\Xsrc$.
Its conditional law is $\Normal((1-t)\Xsrc,t^2I)$, and \cref{eq:forward-path} is its exact Gaussian transition for $\tnext<t<1$, extended to $t=1$ by $k=0$, $s=1$.
For $0<\tnext<t<1$, Gaussian conditioning gives variance $\varsigma^2=\tnext^2s^2/t^2$ and mean
\begin{equation}
\E[Z_{\tnext}^\src\mid Z_t^\src,\Xsrc=x]
=\frac{\tnext^2s^2}{t^2}\Big[\frac{(1-\tnext)x}{\tnext^2}+\frac{kZ_t^\src}{s^2}\Big]
=b_tx+a_tZ_t^\src,
\label{eq:supp-bridge}
\end{equation}
giving \cref{eq:posterior-coefficients}.
The endpoint coefficients are $(a_t,b_t,\varsigma)=(0,1,0)$ at $\tnext=0$ and $(0,1-\tnext,\tnext)$ at $t=1$.
Writing $Z_{\tnext}^\src=a_tZ_t^\src+b_t\Xsrc+\varsigma\eta_t$, with $\eta_t\mid(Z_t^\src,\Xsrc)\sim\Normal(0,I)$, gives $r_t=b_t(\Xsrc-\hat x_t^\src)+\varsigma\eta_t$: source prediction error plus bridge noise.

\paragraph{Connection to synchronous reverse dynamics.}
For $h=t-\tnext\to0$, uniformly for $t$ in compact subintervals of $(0,1)$,
\begin{equation}
a_t=1-\frac{2h}{t}-\frac{h}{1-t}+O(h^2),\qquad
b_t=\frac{2h}{t}+O(h^2),\qquad
s^2,\ \varsigma^2=g(t)^2h+O(h^2).
\label{eq:supp-smallstep}
\end{equation}
Thus $\Phi_t^c(z)-z=h f_c^{\mathrm{rev}}(z,t)+O(h^2)$.
Using \cref{eq:supp-split}, the drift and, for exact models, the score are
\begin{align}
f_c^{\mathrm{rev}}(z,t)&=\frac{2\hat x(z,t,c)}{t}-\frac{(2-t)z}{t(1-t)}=-2v(z,t,c)-\frac{z}{1-t},
\label{eq:supp-drift}\\
\nabla\log p_{c,t}(z)&=-\frac{z-(1-t)\hat x(z,t,c)}{t^2}=-\frac{\hat\varepsilon(z,t,c)}{t}.
\label{eq:supp-tweedie}
\end{align}
For exact models, \cref{eq:supp-drift,eq:supp-tweedie} give $f_c^{\mathrm{rev}}(z,t)=-f^{\mathrm{fw}}(z,t)+g(t)^2\nabla\log p_{c,t}(z)$, where $f^{\mathrm{fw}}(z,t)=-z/(1-t)$.
This identifies the posterior step's limiting drift with the Anderson reverse drift~\cite{anderson1982reverse}.
For $0<\tau<1$ and reverse time $u=\tau-t$, time reversal as used in Sync-SDE~\cite[Sec.~3.3]{zhang2026syncsde} gives, when $\Xsrc\sim p_c$,
\begin{equation}
d\bar Z_u^\src=f_c^{\mathrm{rev}}(\bar Z_u^\src,\tau-u)\,du+g(\tau-u)\,dB_u,
\label{eq:supp-reverse}
\end{equation}
where $\bar Z_u^\src=Z_{\tau-u}^\src$ and $B$ is Brownian for $\bar{\mathcal F}_u=\sigma(Z_s^\src:\tau-u\le s\le\tau)$, with increments independent of $Z_\tau^\src$.
Set $f_\src^{\mathrm{rev}}=f_{\csrc}^{\mathrm{rev}}$, $f_\tar^{\mathrm{rev}}=f_{\ctar}^{\mathrm{rev}}$, and
\begin{equation}
R_u=\bar Z_u^\src-\bar Z_0^\src-\int_0^u f_\src^{\mathrm{rev}}(\bar Z_r^\src,\tau-r)\,dr.
\label{eq:supp-R}
\end{equation}
By \cref{eq:supp-smallstep}, $r_t=Z_{\tnext}^\src-Z_t^\src-hf_\src^{\mathrm{rev}}(Z_t^\src,t)+O(h^2)$.
The target step is therefore first-order consistent with
\begin{equation}
d\bar Z_u^\tar=f_\tar^{\mathrm{rev}}(\bar Z_u^\tar,\tau-u)\,du+dR_u,\qquad
\bar Z_0^\tar=Z_\tau^\src,
\label{eq:supp-target-sde}
\end{equation}
where $\bar Z_u^\tar=\bar Z_u^\src+D_{\tau-u}$.
The results below concern this continuous equation and assume an adapted solution; convergence of the discrete updates is not proved.

\begin{proposition}[Forward-path transport]\label{prop:sync}
Let $\psrc$ and $\ptar$ have finite second moments, let $\Xsrc\sim\psrc$, and let both clean estimates be exact.
Assume that diffusion time reversal holds for \cref{eq:supp-forward} started from $\psrc$ and from $\ptar$, and that the reverse equations \cref{eq:supp-reverse} have, from every initial state at time $\tau$, weak solutions that are unique in law, do not explode, and have a well-defined endpoint kernel at $t=0$.
Consider the forward path with the posterior step on all tokens, without spatial mixing or source anchoring, started at $0<\tau<1$ with $D_\tau=0$, in its continuous-time form \cref{eq:supp-target-sde}.
On the probability space of the source path, assume that \cref{eq:supp-target-sde} admits a nonexplosive continuous solution adapted to $\bar{\mathcal F}$, with an almost-sure limit as $u\uparrow\tau$.
\emph{(i)}~The target copy follows the target reverse SDE from $Z_\tau^\src$, driven by a Brownian motion independent of $Z_\tau^\src$.
\emph{(ii)}~Its output law converges in total variation to $\ptar$ as $\tau\uparrow1$.
\end{proposition}

\begin{proof}[Proof of \cref{prop:sync}]
Part (i) is the synchronous-coupling argument of Sync-SDE~\cite[Sec.~5]{zhang2026syncsde}: \cref{eq:supp-reverse,eq:supp-R} give $dR_u=g(\tau-u)dB_u$, which drives the assumed target solution.
For (ii), let $K_\tar$ be the target endpoint kernel. Uniqueness and time reversal give $\nu_\tau=p_{\src,\tau}K_\tar$ and $p_{\tar,\tau}K_\tar=\ptar$; contraction of total variation yields
\begin{equation}
\|\nu_\tau-\ptar\|_{\mathrm{TV}}
\le \|p_{\src,\tau}-p_{\tar,\tau}\|_{\mathrm{TV}}.
\label{eq:supp-sync-tv}
\end{equation}
As $p_{c,\tau}$ is the mixture over $x\sim p_c$ of $\Normal\big((1-\tau)x,\tau^2I\big)$, convexity of the Kullback--Leibler divergence gives
\begin{equation}
\mathrm{KL}\!\left(p_{c,\tau}\,\|\,\Normal(0,\tau^2I)\right)
\le \frac{(1-\tau)^2}{2\tau^2}\E_{p_c}\|x\|^2.
\label{eq:supp-sync-kl}
\end{equation}
By the triangle inequality and Pinsker's inequality $\|p-q\|_{\mathrm{TV}}\le\sqrt{\mathrm{KL}(p\|q)/2}$, the right side of \cref{eq:supp-sync-tv} is at most $\frac{1-\tau}{2\tau}\big(\sqrt{\E_{\psrc}\|x\|^2}+\sqrt{\E_{\ptar}\|x\|^2}\big)\to0$ as $\tau\uparrow1$, which proves (ii) without a stochastic integral at the singular endpoint $t=1$.
\end{proof}
\Cref{prop:sync} concerns draws $\Xsrc\sim\psrc$; conditional on a fixed source, $R$ need not be Brownian.

\begin{proposition}[Partial increments keep the law]
\label{prop:kappa}
Let $\kappa$ be a per-token weight with values in $[0,1]$ that may depend on the states of the two copies up to the current step, but not on $\Xsrc$ itself, and let the target copy take the step
\begin{equation}
Z_{\tnext}^\tar=b_t\hat x^{\tar}_t+a_tZ_t^\tar+\kappa\odot r_t+\sqrt{1-\kappa^2}\odot\varsigma\,\zeta_t,
\label{eq:supp-kappa}
\end{equation}
with $\zeta_t\sim\Normal(0,I)$ independent of everything else and $\odot$ the elementwise product.
In continuous time this is $d\bar Z_u^\tar=f_\tar^{\mathrm{rev}}(\bar Z_u^\tar,\tau-u)\,du+\kappa\odot dR_u+\sqrt{1-\kappa^2}\odot g(\tau-u)\,dB'_u$, with a Brownian motion $B'$ independent of $Z^\src$, and with $\kappa\in[0,1]$ predictable for the filtration $\mathcal G_u$ of the source path and $B'$ (generated by $\bar{\mathcal F}_u$ and $(B'_r)_{r\le u}$).
For each choice of $\kappa$ under consideration, assume that this equation with $\bar Z_0^\tar=Z_\tau^\src$ admits a nonexplosive $\mathcal G$-adapted continuous solution with an almost-sure limit as $u\uparrow\tau$.
Under the assumptions of \cref{prop:sync}, its conclusions hold for every such solution.
$\kappa\equiv1$ is the posterior step, and $\kappa\equiv0$ uses independent target noise from the shared start $Z_\tau^\src$.
\end{proposition}
\begin{proof}
For such a solution, $B$ and $B'$ are independent Brownian motions for $\mathcal G$, and $dR=g\,dB$, so the target is driven by $g\,d\tilde B$ with $d\tilde B=\kappa\odot dB+\sqrt{1-\kappa^2}\odot dB'$.
$\tilde B$ is a continuous local martingale for $\mathcal G$ with quadratic covariation $d\langle\tilde B_i,\tilde B_j\rangle_u=\delta_{ij}(\kappa_i^2+1-\kappa_i^2)\,du=\delta_{ij}\,du$, hence, by L\'evy's characterization, a Brownian motion for $\mathcal G$, with increments independent of the $\mathcal G_0$-measurable $Z_\tau^\src$.
The proof of \cref{prop:sync} applies with $\tilde B$ in place of $B$.
\end{proof}
Since $\kappa=0$ has the same limiting output law, this marginal guarantee alone does not establish source preservation.

\subsection{Fresh noise averages}
\label{supp:proofs-average}

Fix $\Xsrc$, and for $t\in(0,1]$ and $D,e\in\R^d$, with $d$ the latent dimension, let
\begin{equation}
F_t(D,e)=\hat x\big((1-t)\Xsrc+te+D,\,t,\,\ctar\big)-\hat x\big((1-t)\Xsrc+te,\,t,\,\csrc\big).
\label{eq:supp-F}
\end{equation}
Under fresh noise, each step draws a new $\varepsilon_t\sim\Normal(0,I)$, so $\Delta\hat x_t=F_t(D_t,\varepsilon_t)$; its average over the reference noise at $D_t=D$ is $\overline{\Delta\hat x}_t(D)=\E\,F_t(D,\varepsilon)$, $\varepsilon\sim\Normal(0,I)$.

\begin{proposition}[Fresh noise averages]
\label{prop:average}
Fix a time $t_{\min}>0$, let both clean estimates be jointly continuous in $(z,t)\in\R^d\times[t_{\min},1]$ and $L$-Lipschitz in $z$, and let all tokens start at a grid time $\tau$.
For every source, under fresh noise and the Euler step, $D_t$ converges in $L^2$, uniformly over the grid times in $[t_{\min},\tau]$ as the largest step tends to $0$, to the solution of $\frac{d}{dt}\bar D_t=\big(\bar D_t-\overline{\Delta\hat x}_t(\bar D_t)\big)/t$, $\bar D_\tau=0$, a function of $\Xsrc$ alone; equivalently $\frac{d}{dt}\bar D_t=\E_\varepsilon[V_t^\tar-V_t^\src]$ with the velocities $V_t^\src=v(Z_t^\src,t,\csrc)$ and $V_t^\tar=v(Z_t^\tar,t,\ctar)$ of \cref{eq:prelim-inversion}, since $v=(z-\hat x)/t$.
\end{proposition}
\noindent
$\bar D$ also depends on the prompts and $\tau$, but not on the noise; we do not verify the hypotheses for FLUX.

\par\vskip 0pt plus 6\baselineskip\penalty-150\vskip 0pt plus -6\baselineskip
\begin{lemma}[The update and its average]
\label{lem:supp-F}
Under the hypotheses of \cref{prop:average}, for $t\in[t_{\min},1]$:
\emph{(a)}~$\|F_t(D,e)-F_t(D',e)\|\le L\|D-D'\|$ and $\|F_t(D,e)-F_t(D,e')\|\le2Lt\,\|e-e'\|$;
\emph{(b)}~$\overline{\Delta\hat x}$ is continuous on $[t_{\min},1]\times\R^d$ and $L$-Lipschitz in $D$;
\emph{(c)}~$\E\big\|F_t(D,\varepsilon)-\overline{\Delta\hat x}_t(D)\big\|^2\le4dL^2t^2$.
\end{lemma}
\begin{proof}
\emph{(a)}~Only the target term depends on $D$; $e$ enters both terms with the factor $t$.
\emph{(b)}~By (a), $\|F_t(D,e)\|\le B_0+L\|D\|+2L\|e\|$ with $B_0=\max_{t\in[t_{\min},1]}\|F_t(0,0)\|<\infty$; this bound is integrable in $e$, so dominated convergence gives continuity in $t$, and (a) gives the Lipschitz bound in $D$, uniform in $t$, hence joint continuity.
\emph{(c)}~The mean minimizes the expected squared distance to a constant, so by (a) the left side is at most $\E\|F_t(D,\varepsilon)-F_t(D,0)\|^2\le(2Lt)^2\,\E\|\varepsilon\|^2=4dL^2t^2$.
\end{proof}

\paragraph{Step 1: averaging.}
On the grid $\tau=t^{(0)}>t^{(1)}>\dots$, with $h_k=t^{(k)}-t^{(k+1)}$, $h=\max_kh_k$, and the Euler step $a_k=t^{(k+1)}/t^{(k)}$, $b_k=1-a_k=h_k/t^{(k)}$, independent draws $\varepsilon_k\sim\Normal(0,I)$ give the displacement and its averaged counterpart
\begin{equation}
D^{(k+1)}=a_kD^{(k)}+b_k\,F_{t^{(k)}}(D^{(k)},\varepsilon_k),\qquad
\bar D^{(k+1)}=a_k\bar D^{(k)}+b_k\,\overline{\Delta\hat x}_{t^{(k)}}(\bar D^{(k)}),
\label{eq:supp-two-rec}
\end{equation}
with $D^{(0)}=\bar D^{(0)}=0$.
We show that, whenever $t^{(k)}\ge t_{\min}$,
\begin{equation}
\E\|D^{(k)}-\bar D^{(k)}\|^2\le4dL^2\Big(\frac{\tau}{t^{(k)}}\Big)^{2L}(\tau-t^{(k)})\,h\le\frac{4dL^2}{t_{\min}^{2L}}\,h .
\label{eq:supp-avg-var}
\end{equation}
The noise in each step has mean zero given the past, so errors add up in variance, not in size.
Given $\varepsilon_0,\dots,\varepsilon_{k-1}$, $D^{(k)}$ is fixed and, by the growth bound in (b), square integrable; by (c), $\xi_k=F_{t^{(k)}}(D^{(k)},\varepsilon_k)-\overline{\Delta\hat x}_{t^{(k)}}(D^{(k)})$ has conditional mean zero and conditional second moment at most $4dL^2(t^{(k)})^2$.
The error $\rho_k=D^{(k)}-\bar D^{(k)}$ satisfies $\rho_{k+1}=u_k+b_k\xi_k$, where $u_k=a_k\rho_k+b_k\big(\overline{\Delta\hat x}_{t^{(k)}}(D^{(k)})-\overline{\Delta\hat x}_{t^{(k)}}(\bar D^{(k)})\big)$ is fixed given the past and $\|u_k\|\le(1+b_kL)\|\rho_k\|$ by (b).
The cross term has mean zero and $b_kt^{(k)}=h_k$, so $\E\|\rho_{k+1}\|^2\le(1+b_kL)^2\,\E\|\rho_k\|^2+4dL^2h_k^2$.
Unrolling from $\rho_0=0$ with $(1+b_iL)^2\le e^{2Lb_i}$ and $b_i\le\log(t^{(i)}/t^{(i+1)})$ bounds each product of factors by $(\tau/t^{(k)})^{2L}$, and $\sum_{j<k}h_j^2\le h\,(\tau-t^{(k)})$ gives \cref{eq:supp-avg-var}.

\paragraph{Step 2: Euler's method.}
The averaged recursion is Euler's method for $\frac{d}{dt}\bar D=-G(t,\bar D)$, $G(t,D)=(\overline{\Delta\hat x}_t(D)-D)/t$, whose right side is continuous and $L_G$-Lipschitz in $D$ on $[t_{\min},\tau]$, $L_G=(1+L)/t_{\min}$, by (b).
With $\varpi$ the modulus of continuity of $s\mapsto G(s,\bar D(s))$, the standard Euler error bound gives, for the error $\chi_k=\|\bar D^{(k)}-\bar D(t^{(k)})\|$,
\begin{equation}
\chi_k\le(\tau-t_{\min})\,e^{L_G(\tau-t_{\min})}\,\varpi(h)\quad\text{for }t^{(k)}\ge t_{\min}.
\label{eq:supp-euler}
\end{equation}

\begin{proof}[Proof of \cref{prop:average}]
By \cref{eq:supp-avg-var,eq:supp-euler}, $\E\|D^{(k)}-\bar D(t^{(k)})\|^2\le8dL^2h/t_{\min}^{2L}+2\chi_k^2$ at every grid time in $[t_{\min},\tau]$, a bound independent of $k$ that tends to $0$ with $h$.
\end{proof}
This result covers $t\ge t_{\min}>0$, not the clean endpoint or the adaptive coupling of \method.
It does not establish the law of the final output; finite-step outputs can still vary with the seed (\cref{supp:seeds}).

\subsection{Mass and mean}
\label{supp:proofs-leak}

We compare a lower bound on the final change under the forward path with an upper bound on the fresh-noise displacement away from the clean endpoint.

\paragraph{Setting.}
Let $\mathcal S$ be a set of tokens whose content falls into separated alternatives (for example, two arrangements of the leaves behind a subject): disjoint \emph{cells} $A_1,\dots,A_J$ at pairwise distance at least $\gamma>0$ that carry all of the mass of $x_{\mathcal S}$, the part of $x$ on $\mathcal S$, under both prompts, with $\Delta_{\mathcal S}<\infty$ the diameter of $\bigcup_jA_j$.
$P_c(j\mid z)$ is the probability of $x_{0,\mathcal S}\in A_j$ given $z_\tau=z$ under prompt $c$, and $\sig$ is the logistic function.
In the \emph{one-cell setting}, the tokens in $\mathcal S$ are independent of the others under both prompts, both prompts give $x_{0,\mathcal S}$ the same law within each cell, and the target multiplies the prior odds of $A_1$ by $e^{\delta}$, $\delta\ge0$, without changing the ratios between the other cells.

\begin{proposition}[Mass and mean]\label{prop:leak}
Assume exact models, continuous time, and a common start $0<\tau<1$.
\emph{(i)}~Under the assumptions of \cref{prop:sync}, with the forward path and the posterior step, for $\Xsrc\sim\psrc$,
\[
\E\big\|\Xtar_{\mathcal S}-\Xsrc_{\mathcal S}\big\|^2\;\ge\;\gamma^2\,\E\,\mathrm{TV}\big(P_\tar(\cdot\mid Z_\tau^\src),\,P_\src(\cdot\mid Z_\tau^\src)\big).
\]
In the one-cell setting, the right side is at least $\gamma^2(1-e^{-\delta})\,U(\tau)$, where $U(\tau)=\E\big[P_\src(1\mid Z_\tau^\src)\big(1-P_\src(1\mid Z_\tau^\src)\big)\big]$ is the posterior uncertainty of $A_1$ at the start; $U$ does not increase for later starts.
\emph{(ii)}~With fresh noise and the Euler step, suppose the clean estimates satisfy the hypotheses of \cref{prop:average}, their parts on $\mathcal S$ depend only on the tokens in $\mathcal S$, and $\|\overline{\Delta\hat x}_t(0)_{\mathcal S}\|\le\delta_{\mathcal S}$ for $t\in[t_{\min},\tau]$.
Then, for every source, the small-step limit $\bar D$ of \cref{prop:average} satisfies $\|\bar D_{t,\mathcal S}\|\le C\,\delta_{\mathcal S}$ on $[t_{\min},\tau]$, where $C=\big((\tau/t_{\min})^{\bar L-1}-1\big)/(\bar L-1)$ ($C=\log(\tau/t_{\min})$ for $\bar L=1$), and $\bar L\le\sqrt{2/\pi}\,\Delta_{\mathcal S}/t_{\min}$ is the Lipschitz constant of the noise-averaged target estimate on $\mathcal S$ over $[t_{\min},\tau]$.
In the one-cell setting, $\delta_{\mathcal S}\le\delta\Delta_{\mathcal S}/4$.
\end{proposition}
\noindent
The bound in (i) uses only the source and target endpoint posteriors conditional on the shared noised start. It therefore also holds for SDEdit and every coupling covered by \cref{prop:kappa}, not just the full posterior-step coupling.
In the one-cell setting, (i) gives a source-averaged lower bound on final squared change proportional to $1-e^{-\delta}$, while (ii) gives a per-source upper bound on squared displacement proportional to $\delta^2$ on $[t_{\min},\tau]$. These bounds do not compare both methods at the clean endpoint.
Related distinctions between posterior sampling and distortion appear in restoration~\cite{ohayon2023stochastic,blau2018perception}.

\begin{lemma}[Couplings across cells]\label{lem:cells}
Let $\alpha$ and $\beta$ be laws on a set partitioned into cells $A_1,\dots,A_J$, with cell probabilities $\bar\alpha_j=\alpha(A_j)$, $\bar\beta_j=\beta(A_j)$, and let $E_\gamma$ be the set of points of a cell at distance less than $\gamma$ from another cell.
Every pair $(X,Y)$ with $X\sim\alpha$ and $Y\sim\beta$ satisfies $\Pr[X\text{ and }Y\text{ lie in different cells}]\ge\mathrm{TV}(\bar\alpha,\bar\beta)$ and $\E\|X-Y\|^2\ge\gamma^2\big(\mathrm{TV}(\bar\alpha,\bar\beta)-\alpha(E_\gamma)\big)$.
If the cells are at pairwise distance at least $\gamma$, then $\alpha(E_\gamma)=0$.
\end{lemma}
\begin{proof}
$\Pr[X\in A_j,\,Y\in A_j]\le\min(\bar\alpha_j,\bar\beta_j)$, and $\sum_j\min(\bar\alpha_j,\bar\beta_j)=1-\mathrm{TV}(\bar\alpha,\bar\beta)$.
If $X$ and $Y$ lie in different cells and $X\notin E_\gamma$, then $\|X-Y\|\ge\gamma$; this event has probability at least $\mathrm{TV}(\bar\alpha,\bar\beta)-\alpha(E_\gamma)$.
\end{proof}

\begin{proof}[Proof of \cref{prop:leak}(i)]
\emph{Bound.}
Along the forward path $Z_\tau^\src\mid\Xsrc\sim\Normal\big((1-\tau)\Xsrc,\tau^2I\big)$, so given $Z_\tau^\src=y$, $\Xsrc\sim\psrc(x\mid z_\tau=y)$, and by \cref{lem:transfer} $\Xtar\sim\ptar(x\mid z_\tau=y)$ for almost every $y$.
The conditional law of $(\Xsrc,\Xtar)$ is thus a coupling of the two posteriors, whose restrictions to $\mathcal S$ live on $\bigcup_jA_j$ (a posterior is absolutely continuous with respect to its prior) with cell probabilities $P_\src(\cdot\mid y)$ and $P_\tar(\cdot\mid y)$.
\Cref{lem:cells} for almost every $y$ and the expectation over $y$ give the bound.

\emph{One cell.}
With $\mathcal S$ independent of the other tokens and independent noise per token, $P_c(\cdot\mid y)$ depends on $y$ only through $y_{\mathcal S}$.
With equal within-cell laws, each cell has the same likelihood under both prompts, so Bayes' rule multiplies the posterior odds of $A_1$ by $e^\delta$ and scales the other cells by a common factor at most one; hence $\mathrm{TV}=P_\tar(1\mid y)-P_\src(1\mid y)$.
With $P=P_\src(1\mid y)$ and $\ell=\log\frac{P}{1-P}$,
\[
\mathrm{TV}=\sig(\ell+\delta)-\sig(\ell)=\frac{P(1-P)(e^\delta-1)}{1-P+Pe^\delta},
\]
so $1-P+Pe^\delta\le e^\delta$ gives $\mathrm{TV}\ge(1-e^{-\delta})\,P(1-P)$, and $\sig'\le1/4$ gives $\mathrm{TV}\le\delta/4$.

\emph{Start time.}
For $\tau'<\tau$, the forward path reaches $Z_\tau^\src$ from $Z_{\tau'}^\src$ without using $\Xsrc$, so $P_\src(1\mid Z_\tau^\src)=\E[P_\src(1\mid Z_{\tau'}^\src)\mid Z_\tau^\src]$, and Jensen's inequality for the concave $P\mapsto P(1-P)$ gives $U(\tau')\le U(\tau)$.
\end{proof}

\begin{proof}[Proof of \cref{prop:leak}(ii)]
By \cref{prop:average}, it suffices to bound the solution $\bar D$ of $\tfrac{d}{dt}\bar D_t=\big(\bar D_t-\overline{\Delta\hat x}_t(\bar D_t)\big)/t$, $\bar D_\tau=0$.
The part of $F_t(D,\varepsilon)$ on $\mathcal S$ depends only on $(D_{\mathcal S},\varepsilon_{\mathcal S})$, so $\overline{\Delta\hat x}_{\mathcal S}$ depends only on $D_{\mathcal S}$, and $\bar D_{\mathcal S}$ solves its own equation.
In log time $\theta=\log(\tau/t)$, it reads $\tfrac{d}{d\theta}\bar D_{\mathcal S}=\overline{\Delta\hat x}_{\mathcal S}(\bar D)-\bar D_{\mathcal S}$, so $\bar D_{\mathcal S}(\theta)=\int_0^\theta e^{-(\theta-r)}\,\overline{\Delta\hat x}_{\mathcal S}(\bar D(r))\,dr$.
Since $\|\overline{\Delta\hat x}_{\mathcal S}(D)\|\le\delta_{\mathcal S}+\bar L\|D_{\mathcal S}\|$, $n(\theta)=\|\bar D_{\mathcal S}(\theta)\|$ satisfies $e^\theta n(\theta)\le\delta_{\mathcal S}(e^\theta-1)+\bar L\int_0^\theta e^rn(r)\,dr$, and Gronwall's inequality for $e^\theta n(\theta)$ gives
\[
n(\theta)\le\delta_{\mathcal S}\,\frac{e^{(\bar L-1)\theta}-1}{\bar L-1}\qquad(=\delta_{\mathcal S}\,\theta\text{ at }\bar L=1),
\]
which increases in $\theta$ and is at most $C\delta_{\mathcal S}$ for $\theta\le\log(\tau/t_{\min})$.

\emph{The averaged Lipschitz constant.}
The map $D_{\mathcal S}\mapsto\E_\varepsilon\hat x_{\mathcal S}\big((1-t)\Xsrc+t\varepsilon+D,t,\ctar\big)$ is a Gaussian smoothing, at scale $t$, of a function $f$ with values in the convex hull of $\bigcup_jA_j$ (the exact estimate is a posterior mean).
For a unit vector $u$, Gaussian integration by parts (Stein's lemma) gives its directional derivative $\frac1t\E\big[(f(\cdot+t\varepsilon)-w)\langle\varepsilon,u\rangle\big]$ for any constant vector $w$, since $\E\langle\varepsilon,u\rangle=0$; with $w$ in the hull, its norm is at most $\Delta_{\mathcal S}\E|\langle\varepsilon,u\rangle|/t=\sqrt{2/\pi}\,\Delta_{\mathcal S}/t$, independently of $\gamma$.
The source term of $F$ does not depend on $D$, so this bounds $\bar L$.

\emph{Size of $\delta_{\mathcal S}$.}
In the one-cell setting, $\hat x_{\mathcal S}(z,t,c)=\sum_jP^t_c(j\mid z)\,\bar x_j(z)$, with $P^t_c(j\mid z)$ the probability of $A_j$ given $z_t=z$ and $\bar x_j(z)=\E[x_{0,\mathcal S}\mid z_t=z,\,x_{0,\mathcal S}\in A_j]$ the same under both prompts.
The changes of $P^t_c(\cdot\mid z)$ between the prompts sum to zero, and those with $j\ne1$ share a sign, so $\hat x_{\mathcal S}(z,t,\ctar)-\hat x_{\mathcal S}(z,t,\csrc)=\sum_{j\ne1}\big(P^t_\tar(j\mid z)-P^t_\src(j\mid z)\big)(\bar x_j-\bar x_1)$ has norm at most $\mathrm{TV}\cdot\Delta_{\mathcal S}\le\delta\Delta_{\mathcal S}/4$, by (i) at time $t$.
At $D=0$ both estimates are taken at the same input, so the bound holds for $\overline{\Delta\hat x}_t(0)_{\mathcal S}$.
\end{proof}
The order comparison is asymptotic: $C$ is independent of $\delta$ but can grow rapidly as $t_{\min}$ decreases.

\begin{corollary}[Regional coupling]\label{cor:regional}
Let $\mu$ be a fixed deterministic region, and let $\psrc$ and $\ptar$ factor over the tokens inside and outside it.
Consider \cref{eq:method} without source anchoring, with the forward path and posterior step inside $\mu$ from $0<\tau_1<1$, and fresh noise and the Euler step outside from $0<\tau_0<1$.
Assume the hypotheses of \cref{prop:sync} for the inside laws and those of \cref{prop:leak} for the outside laws.
\emph{(a)}~Inside $\mu$, the updates equal those of the global forward-path construction using the same reference noise. Its continuous-time dynamics therefore satisfy \cref{prop:reach} for the restricted laws, with the mixture assumptions required for part (ii).
\emph{(b)}~Outside $\mu$, the small-step displacement obeys \cref{prop:leak}(ii) on $[t_{\min},\tau_0]$. A global forward-path construction from $\tau_1$ instead obeys the final-output lower bound of \cref{prop:leak}(i) there.
\end{corollary}
\begin{proof}
Factorization and independent noising make each region's exact clean estimates depend only on that region.
Restricting \cref{eq:method} to $\mu$ gives the global posterior update, and restricting it to the complement gives the fresh-noise Euler update, with the corresponding initial conditions.
Thus the inside continuous-time equation is \cref{eq:supp-target-sde}, to which \cref{prop:reach} applies, and the outside averaged equation is that of \cref{prop:average}, to which \cref{prop:leak}(ii) applies.
The global forward-path comparison follows directly from \cref{prop:leak}(i).
\end{proof}
The pure-noise start is analyzed through $\tau_1\uparrow1$ as in \cref{prop:sync}.
DiffEditor~\cite{mou2024diffeditor} also localizes stochastic sampling, in an inversion-based editor.
The result here assumes fixed, independent regions; adaptive expansion and cross-region attention in \method are evaluated empirically in \cref{sec:analysis-empirical}.

\section{Proof of \cref{prop:reach}}
\label{supp:when}

\subsection{Posterior transfer and reachability}
\label{supp:when-reach}

We prove \cref{prop:reach} under the regularity and global continuous-time assumptions of \cref{prop:sync}: (i) is \cref{lem:transfer}, and (ii) computes the probability of reaching the new component.

\begin{lemma}[Posterior transfer]
\label{lem:transfer}
Under the assumptions of \cref{prop:sync}, let $\Xsrc\sim\psrc$ with its forward path $Z^\src$.
In continuous time, the target copy of the forward path with the posterior step and start time $\tau$ satisfies, for almost every $y$,
\begin{equation}
\mathrm{Law}\big(\Xtar\mid Z_\tau^\src=y\big)=\ptar\big(x\mid z_\tau=y\big).
\label{eq:supp-transfer}
\end{equation}
The same holds for SDEdit~\cite{meng2022sdedit}, the target's reverse process started at $Z_\tau^\src$ with independent noise.
\end{lemma}
\begin{proof}
By the proof of \cref{prop:sync}, the target copy solves the reverse equation of $\ptar$ from $Z_\tau^\src$, driven by a Brownian motion independent of $Z_\tau^\src$; given $Z_\tau^\src=y$, it is a weak solution from $y$, so by uniqueness in law its output has the law $K_\tar(y,\cdot)$.
The time reversal of the forward process started from $\ptar$ gives $K_\tar(y,\cdot)=\ptar(x\mid z_\tau=y)$ for $p_{\tar,\tau}$-almost every $y$, and $p_{\tar,\tau}$ and $p_{\src,\tau}$ have positive densities (convolutions with $\Normal(0,\tau^2I)$), so the null sets agree.
SDEdit and every $\kappa$ of \cref{prop:kappa} change only the Brownian motion, which is again independent of $Z_\tau^\src$.
\end{proof}
This identifies the conditional output law, not its coupling with the clean source.

\paragraph{Reachability expression.}
In \cref{prop:reach}(ii), let the old and new components be $\psrc=\Normal(m_{\mathrm{old}},\Sigma)$ and $p_{\mathrm{new}}=\Normal(m_{\mathrm{new}},\Sigma)$, with means $m_{\mathrm{old}},m_{\mathrm{new}}$, shared covariance $\Sigma\succ0$, $m=m_{\mathrm{new}}-m_{\mathrm{old}}\ne0$, and $\ptar=\pi_{\mathrm{old}}\psrc+\pi_{\mathrm{new}}p_{\mathrm{new}}$ with $0<\pi_{\mathrm{new}}<1$ and $\pi_{\mathrm{old}}=1-\pi_{\mathrm{new}}$.
In this mixture model, $\Preach(\tau)$ is the probability that the output's component label $C$, drawn from $\ptar(C\mid x)$, is new. This is the idealized structural-change proxy used in \cref{prop:reach}(ii); the Gaussian components can overlap.
Then
\begin{equation}
\Preach(\tau)=\E[\sig(L)],\qquad
L\sim\Normal(\ell-\lambda/2,\lambda),\qquad
\lambda=(1-\tau)^2\,m^\top\big((1-\tau)^2\Sigma+\tau^2I\big)^{-1}m,
\label{eq:preach}
\end{equation}
with the prior log-odds $\ell=\log(\pi_{\mathrm{new}}/\pi_{\mathrm{old}})$ and $\sig$ the logistic function.

\begin{proof}[Proof of \cref{prop:reach}]
Part (i) is \cref{lem:transfer}.
For (ii), let $a=1-\tau$ and $V=a^2\Sigma+\tau^2I$; under the component with mean $m_\bullet$, $z_\tau\sim\Normal(am_\bullet,V)$, so the log-odds of the new component given $z_\tau=z$ is
\begin{equation}
L(z)=\ell+\log\frac{\Normal(z;am_{\mathrm{new}},V)}{\Normal(z;am_{\mathrm{old}},V)}=\ell+a\,m^\top V^{-1}\Big(z-\frac{a(m_{\mathrm{new}}+m_{\mathrm{old}})}{2}\Big).
\label{eq:supp-logodds}
\end{equation}
By (i), the output given $Z_\tau^\src$ follows $\ptar(x\mid z_\tau=Z_\tau^\src)$; as $C$ and $z_\tau$ are independent given $x$, the tower property gives the probability $\ptar(C=\mathrm{new}\mid z_\tau=Z_\tau^\src)=\sig(L(Z_\tau^\src))$ that the output label is new.
For $\Xsrc\sim\psrc$, $Z_\tau^\src\sim\Normal(am_{\mathrm{old}},V)$, so $L(Z_\tau^\src)$ is Gaussian with mean $\ell-\frac{a^2}{2}m^\top V^{-1}m=\ell-\lambda/2$ and variance $a^2m^\top V^{-1}m=\lambda$, which is \cref{eq:preach}.

As $\tau\uparrow1$, $\lambda\to0$, so $\Preach\to\sig(\ell)=\pi_{\mathrm{new}}$.

For monotonicity, $\Preach(\tau)=\E_{\psrc}[\psi(\Lambda_\tau(Z_\tau^\src))]$ with the likelihood ratio $\Lambda_\tau=e^{L-\ell}$ of the components and the concave $\psi(r)=\pi_{\mathrm{new}}r/(\pi_{\mathrm{new}}r+\pi_{\mathrm{old}})$.
For $\tau'<\tau$, $Z_\tau^\src$ follows from $Z_{\tau'}^\src$ by a forward transition shared by both components, and $\psrc$ is the old one, so $\Lambda_\tau(Z_\tau^\src)=\E_{\psrc}[\Lambda_{\tau'}(Z_{\tau'}^\src)\mid Z_\tau^\src]$, and Jensen's inequality gives Blackwell's ordering of experiments:
\begin{equation}
\Preach(\tau')=\E_{\psrc}\Big[\E_{\psrc}\big[\psi(\Lambda_{\tau'}(Z_{\tau'}^\src))\mid Z_\tau^\src\big]\Big]
\le\E_{\psrc}\big[\psi(\Lambda_\tau(Z_\tau^\src))\big]=\Preach(\tau).
\label{eq:supp-blackwell}
\end{equation}
\end{proof}
\Cref{eq:preach} is a forward--backward class-change probability~\cite[App.~B]{biroli2024dynamical}.
The model assumes one start time for the whole latent and no guidance; with guidance, the estimates are no longer posterior means of fixed distributions, so $\ell$ and $m$ are effective parameters, not measured on FLUX.
\supppart{Image experiments: Setup and quantitative results}

\section{Implementation details}
\label{supp:impl}

\paragraph{Region from attention.}
The probes evaluate both prompts on $(1-t)\Xsrc+t\varepsilon$ at $t\in\{0.7,0.5\}$ and read the text--image part of the joint attention in the single-stream blocks: the logits of image-token queries against text-token keys (after rotary embedding), with the softmax taken over the valid text tokens only, averaged over heads, at the selected text tokens.
The words come from a whole-word diff of the prompts after the T5 tokenizer (plurals and possessives matched, function words dropped).
The initial region $\mu_0$ is the union of the subject and deleted-word maps (the added-word map only if both are empty), each normalized by its $99$th percentile, binarized at $\max(\text{Otsu},0.5)$, and dilated by $2$ latent pixels.
For $t>t_r$ the region expands only inside a ring of $3$ latent pixels around itself, where the target copy's subject-and-added-word map exceeds $\max(\text{Otsu},0.6)$.
Because the region comes from the generator's own attention, it needs no separate segmentation model, also covers words that a segmenter has no mask for (an added object, a style), and carries over unchanged to video, sound, and 3D (\cref{supp:video,supp:apps}).

\paragraph{Scaling back at $t_r$.}
$d$ is the root mean square over channels of $\hat x^{\tar}_t-\hat x^{\src}_t$ per latent pixel, and $\bar d$ its median outside $\mu$, capped at $0.25$.
The agreement weight $\omega$ of \cref{sec:method} is $\omega=1$ for $d<2\bar d$, $\omega=0$ for $d>4\bar d$, linear in between, averaged over $3\times3$ latent pixels and restricted to $\mu$.

\paragraph{Other settings.}
Both copies use the plain prompts at distilled guidance $3.5$ on $16$ steps of the resolution-shifted FLUX schedule; a token outside $\mu$ starts at the first step with $t\le t_0=0.919$, which on FlowEdit's $28$-step grid at $512^2$ is its released start~\cite{kulikov2025flowedit}, and on our $16$-step grid is $t=0.89$; the main text writes this threshold as $t_0=0.92$.
Keys and values are shared, with SynPS's adaptive rotary weight~\cite{chen2026synps}, in the $13$ blocks of FreeFlux~\cite{wei2025freeflux}: $0$, $7$--$10$, $18$, $25$, $28$, $37$, $42$, $45$, $50$, and $56$, counting the $19$ double-stream blocks first.

\section{The SPRITE benchmark}
\label{supp:bench}

We will publicly release all of the source images, editing instructions, and evaluation annotations. All hyperparameters were fixed before evaluation on \sprite. None of the images used to develop our method appear in \sprite.

\paragraph{Images.}
\sprite has 491 edits of 207 images: 166 photographs of Open Images~\cite{kuznetsova2020openimages} (version 7) (validation and test), 2 further Flickr photographs, and the 39 images of previous work's figures~\cite{kulikov2025flowedit, chen2026synps}. Although the authors of SynPS~\cite{chen2026synps} stated that they would release their benchmark, it was not publicly available at the time of writing. We therefore used only the images shown in their paper and had Claude Opus 5.5 write the source and target prompts for each image. We collected 77 images from previous work in total and developed our method on 38 images that are not in SPRITE.
The photographs were screened automatically by Claude Opus 5.5 and reviewed by hand: the subject is in focus, fills $10$--$70\%$ of the frame, is not heavily occluded, and is evenly colored, with no watermark and no public figure; each is cropped to a square around the subject and resized to $512^2$.

\paragraph{Edits.}
We used Claude Opus 5.5 to write, for each edit, two prompts that share one detailed description of the image and differ only in the edited words, which specify a concrete visual target. Each edit also includes a subject phrase, a one-sentence instruction, check questions whose answer is \emph{yes} for a successful edit and \emph{no} for the source, and keep questions about what must not change.
Two reviewers inspected the output and removed low-quality photos and photos with unclear subjects, as well as edits whose targets are implausible or hard to see at $512\times512$ resolution. The manual screening of SPRITE was completed without looking at the output of any method. \cref{tab:supp-cat} reports the edit success rate per category.

\begin{table*}[t]
\centering
\caption{\textbf{Edit success per category on \sprite.}
\method makes more of the edits than every baseline in every category; the gap is largest for shape and state edits, where the layout must change.
Edit as in \cref{tab:main}; settings as there.}
\label{tab:supp-cat}
\small
\setlength{\tabcolsep}{5pt}
\begin{tabular}{lr cccc >{\columncolor{oursrow}}c}
\toprule
category & edits & RF-Solver-Edit & FireFlow & FlowEdit & Sync-SDE & \textbf{\method (Ours)} \\
\midrule
human pose & 89 & \underline{68.5} & 59.6 & \underline{68.5} & 56.2 & \textbf{89.9} \\
animal pose & 78 & \underline{62.8} & 51.3 & 61.5 & 42.3 & \textbf{80.8} \\
interaction & 77 & 42.9 & 33.8 & \underline{48.1} & 32.5 & \textbf{74.0} \\
shape & 71 & 14.1 & 14.1 & 15.5 & \underline{16.9} & \textbf{63.4} \\
state & 54 & \underline{20.4} & 18.5 & 16.7 & \underline{20.4} & \textbf{63.0} \\
expression & 30 & 56.7 & 56.7 & \underline{66.7} & 53.3 & \textbf{73.3} \\
rigid & 92 & 63.0 & 64.1 & \underline{65.2} & 56.5 & \textbf{89.1} \\
\midrule
all & 491 & 48.7 & 43.8 & \underline{50.1} & 40.5 & \textbf{78.0} \\
\bottomrule
\end{tabular}
\end{table*}

\section{Metrics}
\label{supp:metrics}
Image metrics are defined here and video metrics in \cref{supp:video}.
Edit (Qwen, VQA, GPT-5.5, Claude Opus 5), Comp., Id., and Art. are in percent; BG is PSNR in dB and BG L1 in gray levels (0–255).

\paragraph{Edit success.}
On \sprite, \emph{Edit} averages the probabilities of Qwen2.5-VL~\cite{bai2025qwen25vl} 7B and 32B that the check holds and the edit is done, and thresholds the mean at $0.5$.
On PIE-Bench, \emph{VQA} is Qwen2.5-VL-7B's probability that the output shows the target prompt.
GPT-5.5~\cite{openai2026gpt55} and Claude Opus 5~\cite{anthropic2026opus5} are each shown the source and the output, the instruction, and the benchmark's check question (on PIE-Bench, whether the output shows the target prompt).

\paragraph{Identity and artifacts.}
\emph{Id} is Qwen2.5-VL-7B's probability that the subject is the same individual, with pose ignored, and \emph{Art.}\ is the 32B judge's probability of a visible artifact.
\emph{Comp.}\ counts an edit that passes Edit, has an identity probability above $0.5$, and an artifact probability below $0.5$.
Comp.\ and Id are computed over the 461 edits that keep their subject (all but the delete and swap edits).

\paragraph{Background.}
On \sprite, \emph{BG} is the PSNR outside the union of the CLIPSeg~\cite{luddecke2022clipseg} subject masks of the source and the output, dilated by $16$ pixels, so that a subject that moves or changes shape is excluded where it was and where it is.
On PIE-Bench it is measured outside the edit mask, over the $404$ local edits whose source does not already show the target (VQA${}<50$).
\emph{BG L1} is the mean absolute error (0--255) outside the subject.

\section{Ablations}
\label{supp:design}
The rows of \cref{tab:design,tab:anchor} compare regional configurations and component removals on all $491$ \sprite edits; the outside-region forward row changes both its path and matched update.
Differences are against \method on the same edits.

\emph{Noise and step inside the region.}
With the posterior step, fresh noise edits $4.7$ points less than the forward path and has a lower BG L1 ($-2.00$); independent noise ($\kappa\equiv0$) edits $5.5$ less, with a $3.83$\,dB lower BG and a $7.7$ points lower Id.\.
The posterior step edits more than the Euler step for the forward path ($+4.3$) and fresh noise ($+9.6$); for both references, the Euler step keeps more background (lower BG L1).
At the Euler step, the forward path edits $10.0$ points more than fresh noise.

\emph{Noise outside the region.}
The forward path with the posterior step there edits as often ($-1.4$ points), with a $1.8$\,dB lower BG.

\emph{Source anchoring.}
Without scaling back, edit success does not change significantly ($-1.2$ points) and BG is $1.01$\,dB lower.
Without shared keys and values, edit success is unchanged ($+0.2$ points) and Id.\ is $3.3$ points lower; fresh noise with the Euler step inside the region then edits $6.1$ points more than with them and still $8.4$ less than the forward path.

\emph{Probe time and scaling-back time.}
\Cref{tab:probe} changes the time of the attention probes and \cref{tab:return} the time of scaling back, everything else fixed.

\begin{table}[t]
\centering
\caption{\textbf{Time of the attention probes.} The default two probe times edit most; a single probe time changes edit success by at most $2.9$ points and the background L1 by at most $0.8$.
A single time takes $2$ probe calls (source and target) instead of $4$ ($36$ NFE). Columns as in \cref{tab:return}.
Settings: \sprite, everything else fixed.}
\label{tab:probe}
\small
\setlength{\tabcolsep}{3.5pt}
\begin{tabular}{lccccc}
\toprule
probe $t$ & Edit\,$\uparrow$ & Id.\,$\uparrow$ & BG\,$\uparrow$ & BG L1\,$\downarrow$ & Art.\,$\downarrow$ \\
\midrule
\ours $0.7$, $0.5$ (default) & 78.0 & 61.8 & 18.19 & 12.78 & 24.9 \\
\oursvar $0.9$ & 77.4 & 61.0 & 17.96 & 13.56 & 24.8 \\
\oursvar $0.7$ & 75.2 & 59.3 & 17.69 & 12.79 & 25.4 \\
\oursvar $0.5$ & 77.2 & 61.4 & 17.98 & 13.19 & 24.7 \\
\oursvar $0.3$ & 76.4 & 60.9 & 17.65 & 13.31 & 24.9 \\
\bottomrule
\end{tabular}
\end{table}

\begin{table}[t]
\centering
\caption{\textbf{Scaling-back time.} Without scaling back the unedited region changes most; moving the step or adding a second one changes edit success by at most $1.8$ points (not significant), and a second step at $0.45$ adds artifacts.
$t_r$ also ends the expansion of $\mu$. Id.\ and BG as in \cref{tab:anchor}; BG L1 in gray levels (0--255).
Settings: \sprite, everything else fixed.}
\label{tab:return}
\small
\setlength{\tabcolsep}{2.6pt}
\begin{tabular}{lcccccc}
\toprule
scaling back & NFE & Edit\,$\uparrow$ & Id.\,$\uparrow$ & BG\,$\uparrow$ & BG L1\,$\downarrow$ & Art.\,$\downarrow$ \\
\midrule
\ours at $t_r=0.7$ (default) & $38$ & 78.0 & 61.8 & 18.19 & 12.78 & 24.9 \\
\oursvar none & $36$ & 76.8 & 60.7 & 17.18 & 15.01 & 24.1 \\
\oursvar at $t_r=0.8$ & $38$ & 77.0 & 61.0 & 17.98 & 13.11 & 24.9 \\
\oursvar at $t_r=0.6$ & $38$ & 77.6 & 61.3 & 18.19 & 12.59 & 26.2 \\
\oursvar at $0.85$ and $0.7$ & $40$ & 76.2 & 60.2 & 18.20 & 11.97 & 24.1 \\
\oursvar at $0.7$ and $0.45$ & $40$ & 76.6 & 60.5 & 18.52 & 11.26 & 28.8 \\
\bottomrule
\end{tabular}
\end{table}

\section{Inversion-based editors}
\label{supp:more-baselines}

\Cref{tab:supp-stableflow} adds Stable Flow~\cite{avrahami2025stableflow}, which shares keys and values from an inverted source, to the comparison of \cref{tab:main}, and \cref{tab:supp-pie-inv} adds RF-Inversion~\cite{rout2025rfinversion} on the standard PIE-Bench metrics.
Stable Flow runs its released code at its defaults: the image latent, scaled by $1.15$, is inverted with the source prompt in $50$ steps, and the source and target are then sampled together in $50$ steps at guidance $1$ and $3$, the target reading the source's keys and values in the release's vital layers ($150$ NFE).
As for every editor, images are edited at $512^2$.
The RF-Inversion numbers are those reported by DNAEdit~\cite{xie2025dnaedit}, which reruns it on PIE-Bench with FLUX at its default settings; DNAEdit's FlowEdit numbers on the same benchmark (PSNR 21.96, LPIPS 112.2, SSIM 83.1) are close to our run of FlowEdit (21.99, 107.4, 84.3).

\begin{table}[t]
\centering
\caption{\textbf{Comparison with Stable Flow.}
\method succeeds on $36.0$ points more of the \sprite edits than Stable Flow, keeps identity and background better, and scores $21.7$ points higher in VQA on PIE-Bench.
Columns as in \cref{tab:main}, judged by Qwen2.5-VL.
Settings: as in \cref{tab:main}; Stable Flow at its released defaults.}
\label{tab:supp-stableflow}
\small
\setlength{\tabcolsep}{2.4pt}
\resizebox{\ifdim\width>\columnwidth\columnwidth\else\width\fi}{!}{%
\begin{tabular}{lc cccc cc}
\toprule
 & & \multicolumn{4}{c}{\sprite} & \multicolumn{2}{c}{PIE-Bench} \\
\cmidrule(lr){3-6}\cmidrule(l){7-8}
 & NFE & Qwen\un{\%}\,$\uparrow$ & Comp.\un{\%}\,$\uparrow$ & Id.\un{\%}\,$\uparrow$ & BG\un{dB}\,$\uparrow$ & VQA\un{\%}\,$\uparrow$ & BG\un{dB}\,$\uparrow$ \\
\midrule
Stable Flow & 150 & 42.0 & 31.0 & 36.6 & 9.17 & 48.3 & 7.28 \\
\midrule
\ours \textbf{\method (Ours)} & 38 & \textbf{78.0} & \textbf{52.7} & \textbf{61.8} & \textbf{18.19} & \textbf{70.0} & \textbf{14.72} \\
\bottomrule
\end{tabular}}
\end{table}

\begin{table}[t]
\centering
\caption{\textbf{Inversion-based editors on PIE-Bench.}
\method has the highest CLIP and VQA scores; Stable Flow keeps the most of the image outside the edit mask because it makes fewer of the edits (VQA $48.3$).
LPIPS $\times10^3$, SSIM $\times10^2$, from the PIE-Bench code; PSNR, LPIPS, and SSIM outside the edit mask, over the $556$ local edits, CLIP over all $700$; VQA as in \cref{tab:main}; --: not reported.
Settings: PIE-Bench~\cite{ju2024pnpinversion} ($700$ edits), FLUX.1-dev, $512^2$; RF-Inversion as reported by DNAEdit~\cite{xie2025dnaedit}.}
\label{tab:supp-pie-inv}
\small
\setlength{\tabcolsep}{2.4pt}
\resizebox{\ifdim\width>\columnwidth\columnwidth\else\width\fi}{!}{%
\begin{tabular}{lc ccc cc c}
\toprule
 & & \multicolumn{3}{c}{unedited region (556)} & \multicolumn{2}{c}{CLIP} & \\
\cmidrule(lr){3-5}\cmidrule(lr){6-7}
Method & NFE & PSNR\,$\uparrow$ & LPIPS\,$\downarrow$ & SSIM\,$\uparrow$ & whole\,$\uparrow$ & edited\,$\uparrow$ & VQA\un{\%}\,$\uparrow$ \\
\midrule
RF-Inversion & 56 & 20.20 & 179.5 & 69.9 & 24.57 & 22.20 & -- \\
Stable Flow & 150 & \textbf{23.26} & \textbf{69.7} & \textbf{88.4} & 24.75 & 21.47 & 48.3 \\
\midrule
\ours \textbf{\method (Ours)} & 38 & 21.54 & 107.4 & 85.1 & \textbf{26.44} & \textbf{23.26} & \textbf{70.0} \\
\bottomrule
\end{tabular}}
\end{table}

\section{Cost per edit}
\label{supp:cost}
\Cref{tab:supp-cost} times every image editor of \cref{tab:main} on the same GPU and the same $10$ \sprite edits.
\begin{table}[t]
\centering
\caption{\textbf{Cost per edit.}
\method takes less time per edit than every released editor except FireFlow, which runs $18$ NFE, at a similar peak memory.
Time: mean seconds from the image file to the edited image; peak: GPU memory allocated during the edit, model weights included ($31.4$--$33.4$\,GiB); \method runs with deterministic kernels, the others as released.
Settings: FLUX.1-dev at $512^2$ on one GPU (an MI250X graphics compute die); $10$ \sprite edits after $2$ warm-up edits.}
\label{tab:supp-cost}
\small
\setlength{\tabcolsep}{3pt}
\resizebox{\ifdim\width>\columnwidth\columnwidth\else\width\fi}{!}{%
\begin{tabular}{lccc}
\toprule
 & NFE & time (s)\,$\downarrow$ & peak (GiB)\,$\downarrow$ \\
\midrule
RF-Solver-Edit & 60 & 19.15 & 32.73 \\
FireFlow & 18 & 6.06 & 32.73 \\
FlowEdit & 48 & 15.17 & 34.42 \\
Sync-SDE & 48 & 15.24 & 32.31 \\
SAM-Flow & 852 & 289.15 & 34.17 \\
\ours \textbf{\method (Ours)} & 38 & 14.02 & 34.27 \\
\bottomrule
\end{tabular}}
\end{table}

\section{User study}
\label{supp:userstudy}
All participants provided informed consent prior to participation.
The trials are edits randomly selected from the image and video benchmarks of \cref{sec:exp}.
Each trial shows the source, the prompts, and the edits of \method and of a baseline as A and B, without method names and with the sides drawn at random; participants choose the edit that makes the requested change (Q1) and the better edit overall (Q2), with a tie option.
Each participant sees the trials in a different random order (images section first, then videos).
The baseline is rotated so that each participant meets each about equally often.
We include three catch trials, each pairing the unedited source with an output that clearly makes the edit.
Of $16$ participants, $15$ chose the edited output in every catch trial they saw; results are over these $15$.
Every participant chose \method in more than half of their trials (15 of 15 on images, 15 of 15 on videos; two-sided sign test, $p<0.001$).

\supppart{Image qualitative results, limitation, and failure cases}

\section{More image comparisons}
\label{supp:qual-nre}
\begin{figure*}[tp]
\centering
\includegraphics[width=\textwidth]{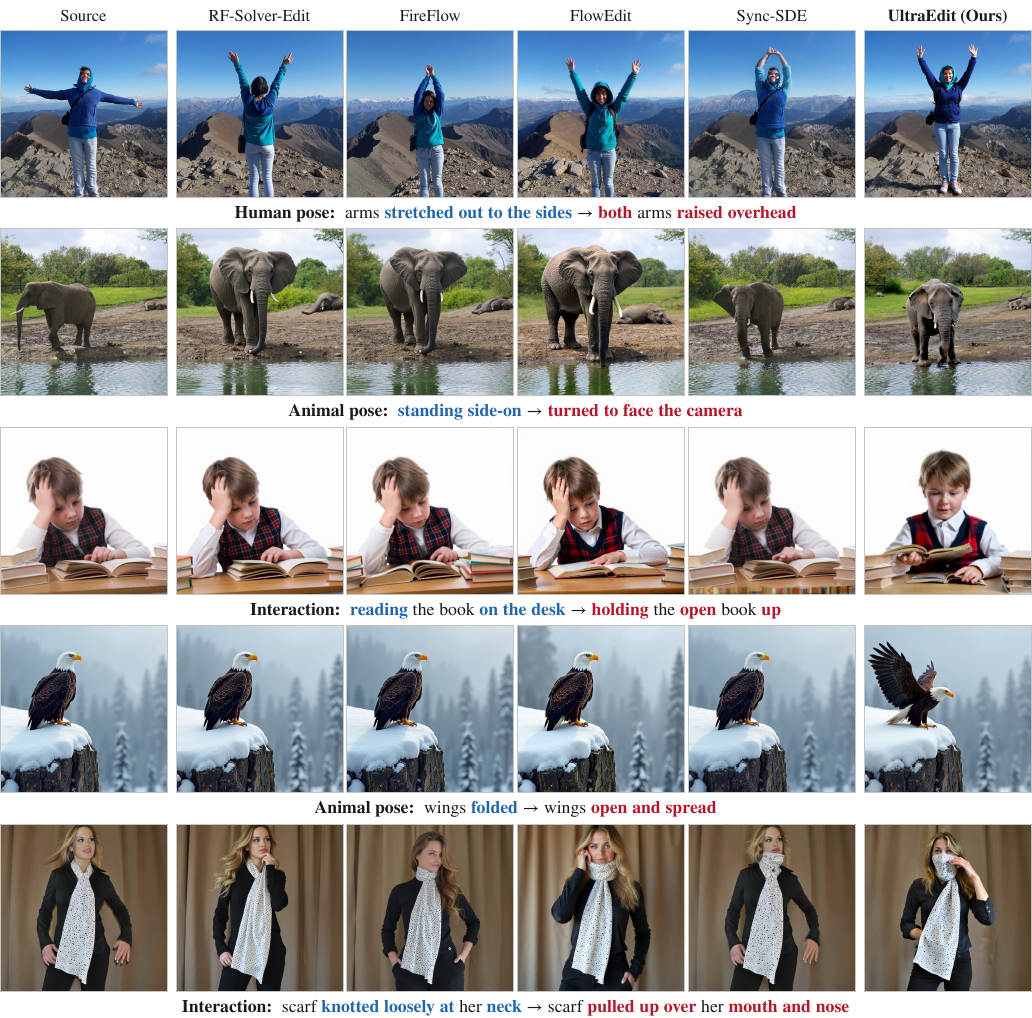}
\caption{\textbf{More comparisons on \sprite.}
Edits we picked as examples (the other five are in \cref{fig:qual}), one output per editor; below each row, the change of the prompt, with the words that change in blue (source) and red (target).}
\label{fig:supp-qual-nre-a}
\end{figure*}

\Cref{fig:qual,fig:supp-qual-nre-a} compare all editors on ten \sprite edits that we picked as examples, one output per editor.
Where the edit moves a part of the subject (the man's hands, the boy's book, the eagle's wings, the panda's legs, the scarf pulled over the mouth) or recolors the suit, the other editors do not make the change and \method does.
On the coffee, RF-Solver-Edit and FireFlow keep it, Sync-SDE makes it paler, FlowEdit turns it olive green, and \method makes a bright green matcha latte.
On the woman's arms and the elephant, RF-Solver-Edit, FireFlow, and FlowEdit succeed as well (on the arms also Sync-SDE), and on the necklace RF-Solver-Edit and FireFlow add green beads; but they often redraw the person or the scene around the edit (the woman turned away, a different face above the necklace, a larger elephant), while \method stays closer to the source.

\section{Zero edits}
\label{supp:zero}

On the eight photos of \cref{fig:preserve}, the whole-image L1 after rounds $1$ to $4$ is 5.68, 6.26, 7.08, and 8.03 for \method; 16.24, 22.44, 26.97, and 30.39 for FireFlow; and 21.73, 31.22, 37.86, and 43.49 for FlowEdit.

\section{Limitations and failure cases}
\label{sec:limitation}
\begin{figure*}[t]
\centering
\includegraphics[width=\textwidth]{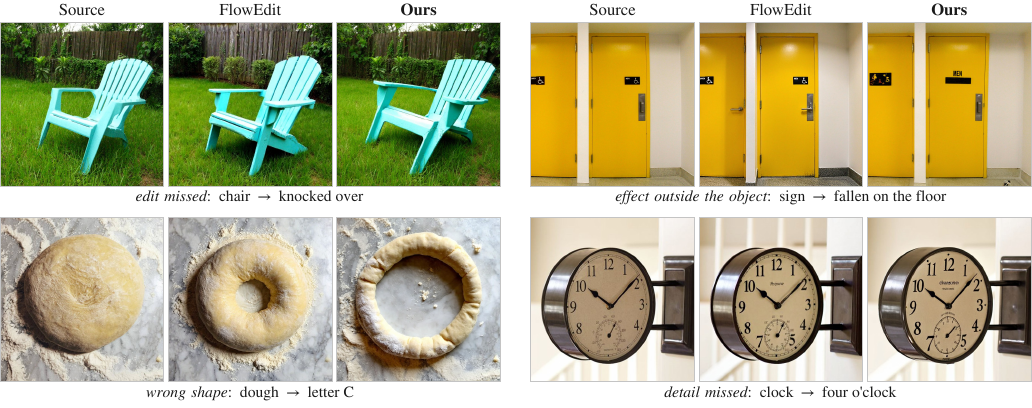}
\caption{\textbf{Failure cases of \method on \sprite.}
The chair is tilted but stays upright; the sign should fall to the floor and is redrawn on the door; the dough becomes a closed ring rather than a C; the clock's hands do not move to four.
Each panel: source, FlowEdit, \method, and the failure type.
One case per failure type, among the edits \method fails.}
\label{fig:failures}
\end{figure*}

\method is training-free and relies on the pretrained generator to turn the target prompt into an image.
While this avoids training, it becomes a limitation when the result must be reasoned out from the prompt.
For a chair to be knocked over onto its side, \method tilts the chair but leaves it standing; FlowEdit also fails on this edit (\cref{fig:failures}). Although attention-based localization can be effective and accurate for structural edits, its reliability may decrease for complex structural changes. In particular, errors in T5 tokenization may lead to inaccurate or incomplete cross-attention maps during probe attention computation, resulting in an incorrectly identified initial editing region. This could limit the method's ability to capture the full spatial extent of complex structural changes.

\paragraph{Failure cases.}
\Cref{fig:failures} shows representative failures of \method on \sprite.
Of the 108 \sprite edits that \method fails, 46 are edits the generator does not render (45 of them also fail for every baseline), 39 are partial edits, 4 come from a wrong region, 15 merge, misplace, or distort the subject, and 4 are correct but judged as failures.

\paragraph{Global edits.}
For a global edit such as a style change, the prompts usually differ only in added words (77 of the 80 PIE-Bench style edits), and $\mu_0$ is their initial attention map.
On these edits, $\mu$ covers a median of 75\% of the image and never all of it, so \method does not reduce to the global forward-path scheme.

\paragraph{Changes outside the subject.}
Shadows, reflections, and contact regions are reached as $\mu$ expands.
On 44 \sprite edits whose result involves them, on average 18.3\% of the pixels that \method changes by more than 40 levels lie outside $\mu_0$, but only 4.7\% lie outside the final $\mu$.

\supppart{Video experiments}

\section{Video editor and benchmarks}
\label{supp:video}

The video editor applies the field of \cref{sec:method} to Wan2.1-T2V-1.3B~\cite{wan2025wan}.

\paragraph{Backbone and grid.}
We run it through Diffusers at $832\times480$, $41$ frames at $16$ frames per second (fps), $17{,}160$ tokens.
Both copies use true classifier-free guidance of scale $5$, with Wan's default negative prompt.
The $16$ steps are uniform in $t$ before Wan's shift of $3$; $t_r=0.7$ is a grid point, and outside $\mu$ the first step after $t_0=0.919$ is at $t=0.900$.

\paragraph{Region.}
The map follows the image editor (\cref{supp:impl}), except as follows.
It reads the cross-attention of video tokens to the umT5 text tokens in layers $2$--$19$ of Wan's $30$ blocks, on the conditional input only; the other layers give diffuse maps.
Each word's map is normalized by its $0.99$ quantile over the whole video, and a word whose normalized map has a median above $0.5$ is dropped as diffuse.
The map of a word set is the maximum over its words, smoothed over $3\times3$ tokens and thresholded at $\max(\text{Otsu},0.5)$, with one Otsu threshold for the whole video.
$M$ is the dilated subject map, blurred by a Gaussian with a standard deviation of $2$ latent pixels, inside $\mu$; the target map that joins it at $t_r$ is blurred too.
All spatial operations act frame by frame.

\paragraph{Noise path and gate.}
Inside $\mu$ the source copy follows the forward path of \cref{eq:forward-path}, outside it fresh noise (\cref{eq:structured,eq:method}); inside $\mu$ the target copy takes the step of \cref{prop:kappa} with $\kappa=c_t$, which for the displacement reads
\begin{equation}
\begin{aligned}
D_{\tnext}={}&b_t\Delta\hat x_t+a_tD_t\\
&+(\kappa-1)\odot r_t+\sqrt{1-\kappa^2}\odot\varsigma\,\zeta_t,
\end{aligned}
\label{eq:supp-video-step}
\end{equation}
with $r_t=Z_{\tnext}^\src-b_t\hat x^{\src}_t-a_tZ_t^\src$ the path's residual and $\zeta_t\sim\Normal(0,I)$ drawn afresh; $\kappa\equiv1$ gives \cref{eq:method}.
The gate measures where the two clean estimates disagree: with $d$ the root mean square over the $16$ channels of $\hat x^{\tar}_t-\hat x^{\src}_t$ at each latent pixel and $\bar d$ its median over the latent pixels with $M<0.5$,
\begin{equation}
\begin{aligned}
c_t={}&\max\Big\{\mathrm{pool}\Big(\mathrm{clip}\Big(\frac{k_2-d/\bar d}{k_2-k_1},0,1\Big)\Big),\ 1-M\Big\},\\
c_t\leftarrow{}&\min\{c_t,\,c_{t^-}\},\qquad (k_1,k_2)=(2,4),
\end{aligned}
\label{eq:supp-gate}
\end{equation}
where pool averages over $3\times3$ latent pixels within each frame and $t^-$ is the previous grid time.
The maximum with $1-M$ releases the gate only inside $M$, the running minimum keeps it from rising again, and at the last step $\kappa=c_t\odot(1-M)$.
Because $c_t$ reads $M$, which the probes compute from the source, its effect is measured (\cref{tab:supp-video-ctrl}) rather than covered by \cref{prop:kappa}.
On the benchmarks, holding $\kappa$ at $1$ leaves edit success unchanged and raises the warping error (\cref{tab:supp-video-ctrl}).

\paragraph{Scaling back and sharing.}
Scaling back is applied once, at the grid point $t=0.700$, with the rule of the image editor (\cref{supp:impl}), its $3\times3$ average taken within each frame, and the step is evaluated again with the same noise.
Wan uses 3D rotary positions and separate self- and cross-attention, and we know no set of its blocks in which keys and values can be shared without their positions; the video editor shares none.

\paragraph{Cost.}
An edit takes $16$ steps of two copies, four probes, and two calls for scaling back: $38$ guided calls, \ie $76$ NFE.
Wan-Edit's released setting ($50$ UniPC steps, the first $15$ skipped, four transformer evaluations per step) takes $140$ NFE.

\paragraph{Design controls.}
\Cref{tab:supp-video-ctrl} changes one choice of the video editor on all IVEBench motion and NRVBench edits.
Fresh noise with the Euler step inside $\mu$ makes $13.6\%$ of the motion edits against $37.3\%$ ($-23.7$ points) and lowers NRVE-Acc by $2.04$: on video, too, the forward path with the posterior step edits more inside the region than fresh noise with the Euler step, a change of noise and step together.
Against holding the gate at $1$, the gate leaves edit success unchanged and lowers the warping error by $1.04$ and $0.88$.
Regenerating the subject's mask edits less (AS $-23.7$ points; NRVE-Acc $-1.40$).

\begin{table}[t]
\centering
\caption{\textbf{Video design controls.}
Inside the region, fresh noise with the Euler step makes fewer edits than the forward path, and the gate lowers the warping error at unchanged edit success.
Rows change one choice of \method (fresh noise with the Euler step: noise and step together), except the last, a design of its own: the source subject's mask (IVEBench: tracked Grounding DINO and SAM~2 mask; NRVBench: its masks), dilated by $16$ pixels, is regenerated from $t=1$ with $\kappa=0$ and every other token is held at the source.
Gate at $1-M$: inside $M$ the target takes its own noise.
Warp: optical-flow self-warping error $\times10^3$ (lower is smoother).
Settings: one run per clip for every row, so on IVEBench \method's AS and IF differ from the means over three runs in \cref{tab:video}; other metrics as there.}
\label{tab:supp-video-ctrl}
\small
\setlength{\tabcolsep}{3pt}
\resizebox{\ifdim\width>\columnwidth\columnwidth\else\width\fi}{!}{%
\begin{tabular}{l ccc}
\toprule
\textbf{IVEBench motion}, 59 edits & AS\,$\uparrow$ & IF\,$\uparrow$ & Warp\,$\downarrow$ \\
\midrule
\ours \textbf{\method (Ours)}, gate $\kappa=c$ & 37.3 & 2.39 & 10.99 \\
\oursvar fresh noise, Euler step in $\mu$ & 13.6 & 1.80 & 10.79 \\
\oursvar gate held at $1$ & 33.9 & 2.31 & 12.03 \\
\oursvar gate held at $1-M$ & 33.9 & 2.17 & 9.83 \\
regenerate the subject's mask, keep the rest & 13.6 & 1.92 & 8.75 \\
\midrule
\textbf{NRVBench}, 180 edits & NRVE-Acc\,$\uparrow$ & $S_\mathrm{instr}$\,$\uparrow$ & Warp\,$\downarrow$ \\
\midrule
\ours \textbf{\method (Ours)}, gate $\kappa=c$ & 34.63 & 73.43 & 7.19 \\
\oursvar fresh noise, Euler step in $\mu$ & 32.60 & 69.26 & 7.33 \\
\oursvar gate held at $1$ & 34.60 & 73.98 & 8.07 \\
\oursvar gate held at $1-M$ & 33.75 & 72.22 & 6.76 \\
regenerate the subject's mask, keep the rest & 33.23 & 69.26 & 5.76 \\
\bottomrule
\end{tabular}}
\end{table}

\paragraph{Benchmarks.}
\emph{FiVE-Bench}~\cite{li2025five} has 419 source--target pairs on $100$ videos ($74$ from DAVIS, $26$ generated; the paper states $420$) in six edit types; its non-rigid swaps keep the source motion.
We use the first $41$ frames, resized to $832\times480$ at $16$\,fps, and its official evaluation (background PSNR and LPIPS, structure distance, CLIP scores, motion fidelity (MFS), and FiVE-Acc, the mean of Qwen2.5-VL-7B's yes/no and multiple-choice accuracy, averaged over the edit types with equal weight).
On Wan-Edit's released outputs our evaluation matches the published numbers (FiVE-Acc $46.46$ against $46.97$, CLIP$_\mathrm{e}$ $21.36$ against $21.23$).
\emph{NRVBench}~\cite{qu2026nrvbench} has one edit on each of its $180$ videos, $30$ in each of six physical categories.
We take the first $41$ native frames, resized to $832\times480$; in the official evaluation Qwen2.5-VL 7B, 32B, and 72B score physical plausibility, temporal coherence, and instruction following, and NRVE-Acc averages the entry score $20\cdot(\text{phys}\cdot\text{temp}\cdot\text{instr})^{1/3}$ over the categories.
Our re-run of Wan-Edit's released code is within 0.04, 0.50, and 0.58 of its published NRVE-Acc for the three judges.
In \cref{tab:video}, Wan-Edit's FiVE-Acc and CLIP$_\mathrm{e}$ on FiVE-Bench and its NRVE-Acc and $S_\mathrm{instr}$ on NRVBench are the published numbers, which our evaluation reproduces as above (NRVBench $34.53$ and $65.83$ against $34.55$ and $65.18$); its GPT-5.5, Claude Opus 5, and BG scores, and all of its IVEBench scores, are our evaluation of its released outputs (FiVE-Bench) or of our run of its released code (NRVBench, IVEBench), and every \method score is our evaluation with the same code.
\emph{IVEBench}~\cite{chen2025ivebench}: we use $59$ of its $64$ short subject-motion entries, kept by visibility and cut rules fixed before any method was run.
We wrote the source and target prompts by the protocol of \sprite and kept IVEBench's instruction for the instruction-based judge; our scores are therefore not comparable with IVEBench's official ones, reported only over its $600$ videos.
Subject masks come from Grounding DINO~\cite{liu2024groundingdino} and Segment Anything Model~2 (SAM~2)~\cite{ravi2025sam2} tracking.

\paragraph{Metrics.}
FiVE-Acc, NRVE-Acc, AS, and the GPT-5.5 and Claude Opus 5 rates are in percent; CLIP$_\mathrm{e}$, FiVE-Bench's CLIP similarity of the edited region, is $\times100$; $S_\mathrm{instr}$, NRVBench's instruction score, is out of $100$; and IF is VEFX-Reward's $1$--$4$ score.
GPT-5.5 and Claude Opus 5 are shown eight frames of the source and of the output, the instruction, and the benchmark's check question.
On all three benchmarks we also compute the following scores.
\emph{Action success} (AS) is judged by Qwen3-VL-32B~\cite{bai2025qwen3vl} on the whole video: it is $1$ if the probability of yes to the edit's check question and the probability of choosing the target over the source description both exceed $0.5$.
On known cases it gave 3 true positives, 0 false positives, 33 true negatives, and 1 false negative; the source videos score 1.9\%, 14.4\%, and 0.0\% on FiVE-Bench, NRVBench, and IVEBench.
\emph{VEFX-Reward-32B}~\cite{gao2026vefx}, with its official code, scores instruction following (IF) from $1$ to $4$.
\emph{Background} follows the rule of \sprite (\cref{supp:metrics}): PSNR and LPIPS outside the union of the source and edited subject masks, dilated by $16$ pixels (the source mask alone for a removal, the edited one alone for an addition).
The masks come from Grounding DINO and SAM~2 tracking rather than per-frame CLIPSeg, so that they follow the subject through every frame, and PSNR pools the background pixels of the whole clip.

\section{More video edits}
\label{supp:video-more}
\begin{figure*}[tp]
\centering
\includegraphics[width=\textwidth]{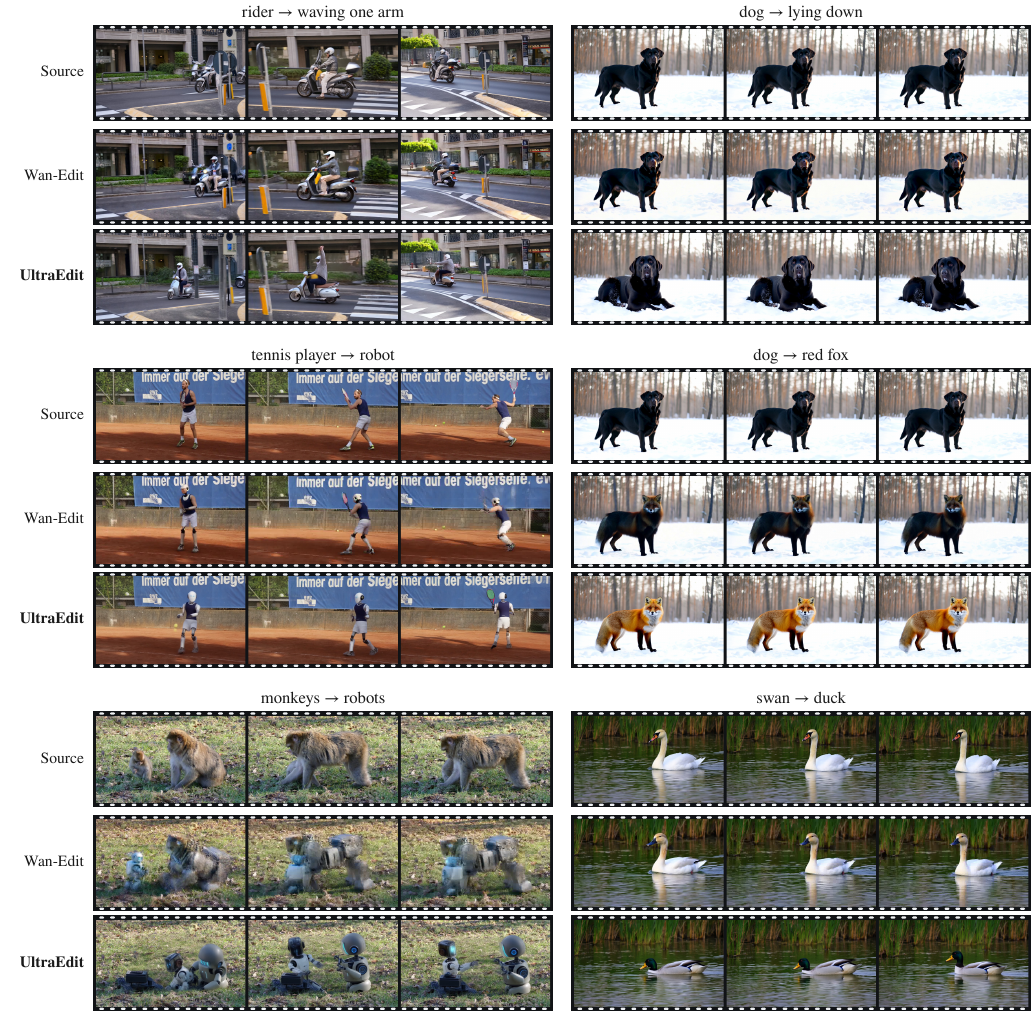}
\caption{\textbf{More video edits (1 of 3): motion, people, and animals.}
\method makes the rider wave and lays the dog down.
Frames $1$, $21$, $41$ of $41$.
Wan-Edit is the output released with FiVE-Bench on its clips and our run of its code on the others.}
\label{fig:supp-video-more-a}
\end{figure*}

\begin{figure*}[tp]
\centering
\includegraphics[width=\textwidth]{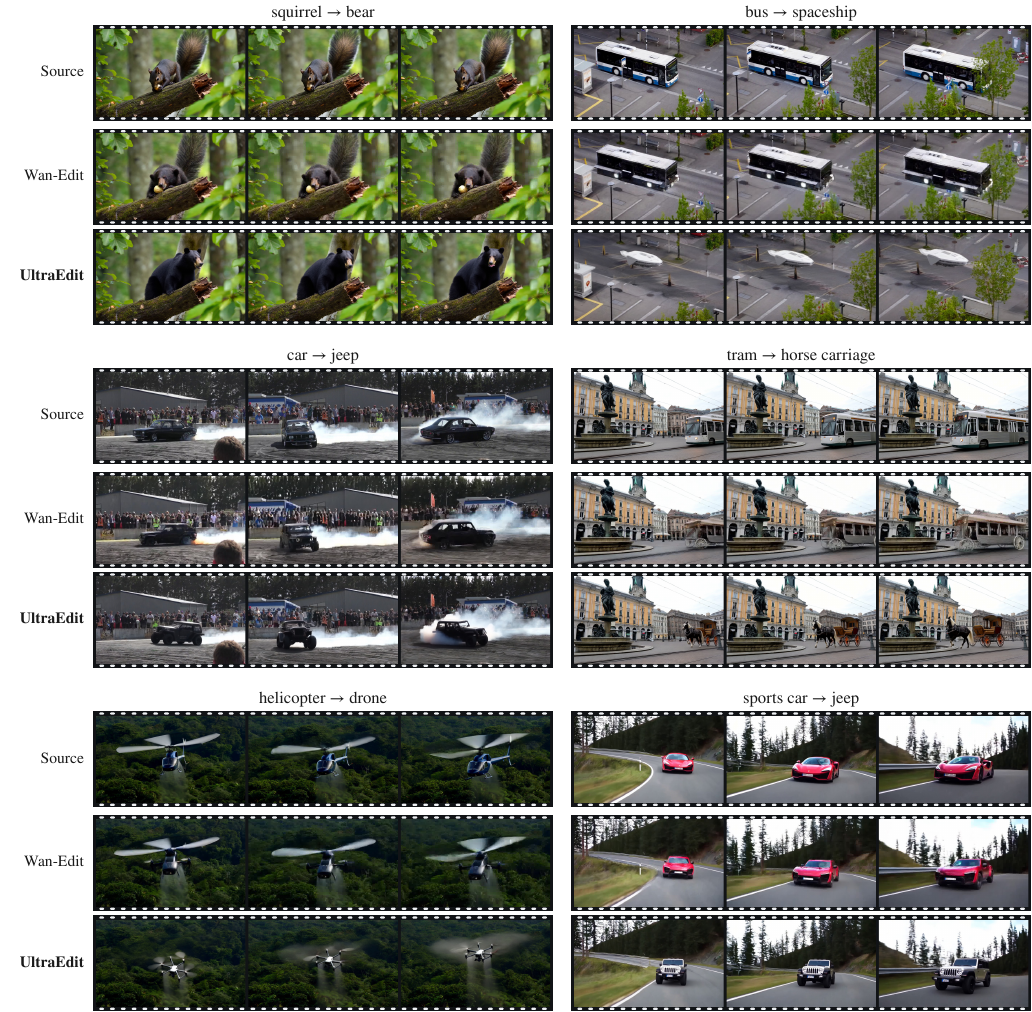}
\caption{\textbf{More video edits (2 of 3): object swaps.}
\method replaces the squirrel with a whole bear, the bus with a hovering spaceship, and the tram with a horse and carriage.
Frames $1$, $21$, $41$ of $41$.
Held-out FiVE-Bench edits; Wan-Edit from the outputs released with the benchmark.}
\label{fig:supp-video-more-b}
\end{figure*}

\begin{figure*}[tp]
\centering
\includegraphics[width=\textwidth]{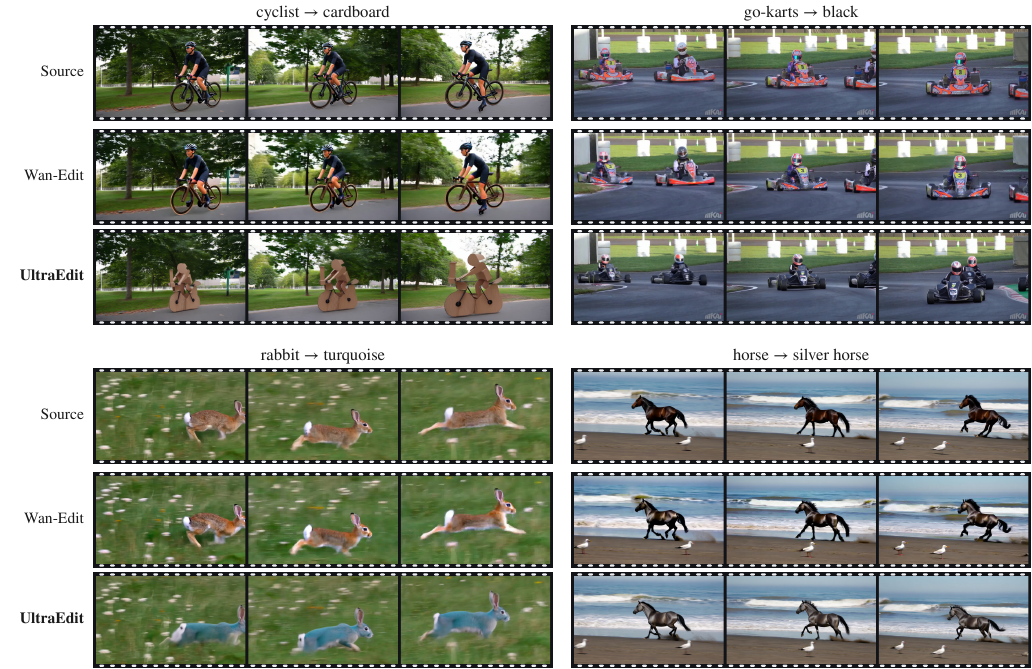}
\caption{\textbf{More video edits (3 of 3): material and color.}
\method replaces the cyclist with a cardboard figure and recolors the go-karts, the rabbit, and the horse.
Frames $1$, $21$, $41$ of $41$.
Held-out FiVE-Bench edits; Wan-Edit from the outputs released with the benchmark.}
\label{fig:supp-video-more-c}
\end{figure*}

\Cref{fig:supp-video-more-a,fig:supp-video-more-b,fig:supp-video-more-c} show sixteen more edits that we picked as examples, with the editors of \cref{fig:video}: thirteen held-out FiVE-Bench edits and three on clips of our development set, which we used to tune the video method.
As in \cref{fig:video}, Wan-Edit mostly keeps the source's outline and changes its texture; it succeeds in part on the burnout car (FiVE's judge also passes its squirrel) and leaves the dog standing.
\method draws the target objects, recolors the go-karts and the rabbit, makes the rider wave, and lays the dog down.

\begin{table}[tp]
\centering
\caption{\textbf{Published video editors} on FiVE-Bench~\cite{li2025five} and NRVBench~\cite{qu2026nrvbench}, as reported by the benchmark papers (not run by us); the last row is \method from \cref{tab:video}. NRVE-Acc: mean of the three Qwen2.5-VL judges of NRVBench's tables.}
\label{tab:supp-video-published}
\providecommand{\vidgray}[1]{#1}
\small
\begin{tabular}{lcc}
\toprule
Method & FiVE-Acc\,$\uparrow$ & NRVE-Acc\,$\uparrow$ \\
\midrule
\vidgray{Pyramid-Edit} & \vidgray{43.84} & \vidgray{33.36} \\
\vidgray{DMT} & \vidgray{48.42} & \vidgray{--} \\
\vidgray{VideoGrain} & \vidgray{37.23} & \vidgray{--} \\
\vidgray{AnyV2V} & \vidgray{38.02} & \vidgray{30.08} \\
\vidgray{TokenFlow} & \vidgray{27.43} & \vidgray{32.56} \\
\vidgray{VidToMe} & \vidgray{26.77} & \vidgray{31.89} \\
\vidgray{VM-Edit} & \vidgray{--} & \vidgray{33.38} \\
\midrule
\ours \textbf{\method (Ours)} & \textbf{53.05} & \textbf{34.63} \\
\bottomrule
\end{tabular}
\end{table}

\supppart{Other generators and tasks}

\section{Other generators and tasks}
\label{supp:apps}

\Cref{tab:supp-ports} lists the ports of \cref{sec:apps}.
Each keeps the update of the image editor: the forward path with the posterior step inside the region $\mu$, fresh noise with the Euler step outside, the two start times, and scaling back.
Two things depend on the generator.
The map $\mu$ is read from what the generator exposes: cross-attention to the edited words on video, sound, and SD3; word timings on speech, whose transformer has no cross-attention; and the source's occupancy on 3D, where cross-attention misses the sparse object.
The weight $\kappa$ of the source increment inside $\mu$ (\cref{eq:supp-kappa}; $\kappa\equiv1$ on images) is, on video, sound, speech, and 3D, the agreement gate $c$ of \cref{eq:supp-gate}, which keeps the increment where the two copies agree; with $\kappa=1$, video showed a higher warping error (\cref{tab:supp-video-ctrl}), sound bursts of noise, and 3D shapes lost their geometry.
Each port runs at its generator's own guidance, classifier-free on all but FLUX, whose guidance is distilled.
Sharing keys and values, part of source anchoring on FLUX, is ported to SD3; the other ports anchor the source by scaling back and the gate.
Except on SD3, $\mu\equiv0$ is the same code with an empty map.

\begin{table}[!h]
\centering
\caption{\textbf{Ports of the update.} Each port uses the displacement step of \cref{eq:offset}: a forward path inside the region $\mu$, fresh noise outside it, and scaling back at $t_r=0.7$. $\kappa$: the agreement gate $c$ of \cref{eq:supp-gate}, or the full increment ($\kappa=1$); $t_0$: start outside $\mu$. Calls: network calls per edit; a guided call batches the conditional and the unconditional input, except on SD3, where the two rows count separately. CFG: classifier-free guidance scale.}
\label{tab:supp-ports}
\small
\setlength{\tabcolsep}{4pt}
\begin{tabular}{@{}lllcccccc c@{}}
\toprule
generator & signal & map $\mu$ & $\kappa$ in $\mu$ & steps & CFG & $t_0$ & calls & calls, $\mu\equiv0$ & edits$\,\times\,$runs \\
\midrule
Wan2.1 (1.3B) & video & attention, per frame & $c$ & 16 & 5 & 0.919 & 38 & -- & Sec.~\ref{supp:video} \\
TangoFlux & sound & attention, time axis & $c$ & 25 & 4.5 & 0.919 & 56 & 44 & $30\times2$ \\
F5-TTS v1 & speech & word alignment & $c$ & 32 & 2 & 0.919 & 66 & 46 & $30\times2$ \\
TRELLIS & 3D & occupancy box & $c$ & 25 & 7.5 & 0.818 & -- & -- & $127\times4$ \\
SD3 Medium & image & attention & $1$ & 16 & 4.5 & 0.85 & 72 & 132 & $41\times1$ \\
\bottomrule
\end{tabular}
\\[2pt]{\footnotesize SD3, calls with $\mu\equiv0$: the released FlowEdit setting for SD3 ($50$ steps), not the same code with an empty map.}
\end{table}

\paragraph{Sound.}
TangoFlux~\cite{hung2024tangoflux}, a $515$M rectified-flow transformer, runs both copies at guidance $4.5$ on $25$ uniform steps.
The map lies on the time axis: probes at $t=0.7$ and $0.5$ read the attention of audio tokens to the edited words in the single-stream blocks; two filters drop diffuse words and words anti-correlated with the source subject words, and the curves are normalized, combined, and thresholded as on video, smoothed over $3$ frames, and dilated by $4$ frames.
Outside $\mu$ the start is $t_0=0.919$; scaling back at $t_r=0.7$ follows the image editor with a $3$-frame average.
On $30$ edits of $10$ clips (two runs each), against $\mu\equiv0$ at the same setting, \method raises the LAION CLAP text score from 0.164 to 0.215 ($p=0.026$).
Tempo and rhythm edits, checked from onsets, pass on 10/14 runs, against 2/14 for $\mu\equiv0$ at the same setting (sign test 8:0, $p=0.008$).
At zero edit, the log-mel L1 to the source differs from that of the VAE reconstruction by at most 0.17\,dB.

\paragraph{Speech.}
F5-TTS v1 Base~\cite{chen2025f5tts}, a $336$M diffusion transformer trained by flow matching, runs $32$ steps with sway sampling ($-1$) and guidance $2$ for both copies.
It has no cross-attention, so $\mu$ is the span of the changed words from wav2vec2 forced alignment, widened by $0.1$\,s.
Outside $\mu$ the start is $0.919$; scaling back caps its floor at $0.15$, and the gate and the last step are those of the video editor.
As in moving to a point (below), the forward path is drawn from the source warped on the time axis to the target timing (the changed span stretched to its new length, the rest shifted), and only the changed span is coupled.
The $30$ edits on $10$ LibriTTS-R test-clean sentences~\cite{koizumi2023librittsr} change words ($22$), timing ($6$, $2$ of them pauses), or the intonation ($2$); each runs twice and passes if an automatic check does.
\method passes 43/44 word, 8/8 rate, 2/4 pause, and 0/4 intonation edits, with log-mel L1 of 1.76 and 2.43\,dB on the unedited prefix and suffix.
$\mu\equiv0$ with the same warp, guidance, and start passes 23/44 word edits and adds no pause (0/4); without the warp it passes 19/44 word edits and no rate or pause edit.
The warp sets the new timing (alone it passes the 8 rate edits) and the coupling synthesizes the changed words (the warp alone passes only the 10 word deletions).

\begin{figure*}[tp]
\centering
\includegraphics[width=\textwidth]{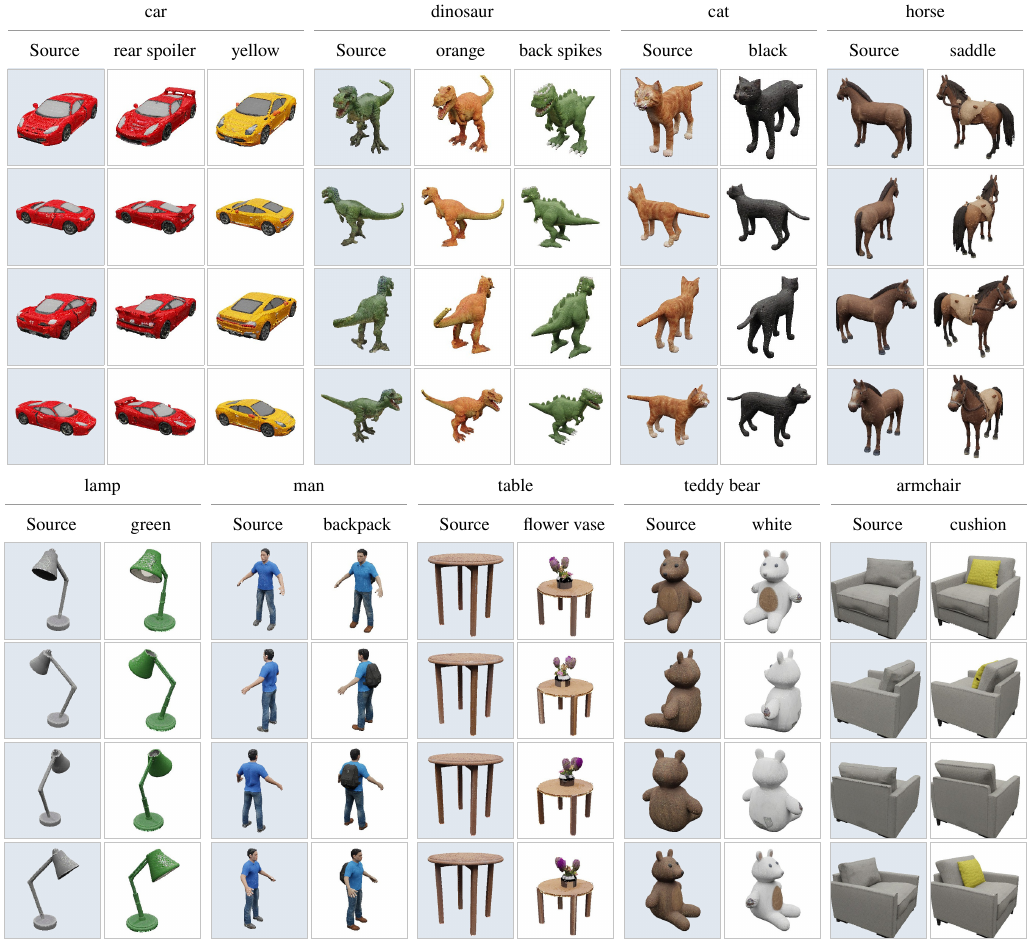}
\caption{\textbf{More 3D edits.}
On TRELLIS assets \method changes colors and adds parts: a spoiler, spikes, a saddle, a backpack, a vase, and a cushion.
Each group shows a source (shaded) and its edits from four sides, $90^\circ$ apart; the top row is the view of \cref{fig:teaser}.
Examples we picked.
Settings as in \cref{supp:apps}; for the car, dinosaur, cat, and armchair, $\mu$ comes from cross-attention, as for six cells of \cref{fig:teaser}.}
\label{fig:supp-3d-more}
\end{figure*}

\paragraph{3D shapes.}
TRELLIS-text-xlarge~\cite{xiang2025trellis} ($2.0$B) generates in two rectified-flow stages, a dense structure stage on $16^3$ tokens and a sparse latent stage on the occupied voxels; both are coupled, and in the sparse stage new voxels have no source and start at $t=1$.
Both copies use guidance $7.5$ inside TRELLIS's guidance interval on its $25$-step grid; outside $\mu$ the start is $0.818$, the one Nano3D~\cite{ye2025nano3d} uses for 3D editing with FlowEdit.
Cross-attention to the subject words misses the object (about $90\%$ of the $16^3$ grid is empty), so $\mu$ is the source's occupancy box enlarged by $1.4$.
The two copies are evaluated in separate calls (batched calls are not bit-identical on our GPUs), the gate's floor is $0.01$, and the last step uses $\kappa=c\,(1-M(1-\omega))$, with $\omega$ the agreement weight at the last step; a zero edit returns the source exactly (intersection over union (IoU) of the occupancy grids $100.0\%$).
The $127$ edits on $37$ assets that TRELLIS generated from our prompts change pose ($66$), parts ($32$), color or material ($22$), and shape ($7$); each runs four times, $P$ is Qwen2.5-VL-7B's probability of yes for the edit on a $2\times2$ grid of views, and $4$ of the $131$ edits we wrote are left out because their check already holds on the source ($P>50\%$).
Over the $127$ edits, $P$ is $53.92\%$ for \method and $45.15\%$ for $\mu\equiv0$ at the same guidance and start $0.957$ ($+8.77$ points, $p=0.005$); $20$ edits succeed in all four runs with \method but not with $\mu\equiv0$, and $7$ the other way round (a lying deer $95\%$ against $24\%$, a taller cottage $86\%$ against $32\%$).
The gate is needed: with $\kappa=1$, subject DINO is 63.2 and IoU 13.9\% (gate: DINO $+6.6$, $p<0.001$); \cref{fig:supp-3d-more} shows further edits.

\begin{figure*}[tp]
\centering
\includegraphics[width=\textwidth]{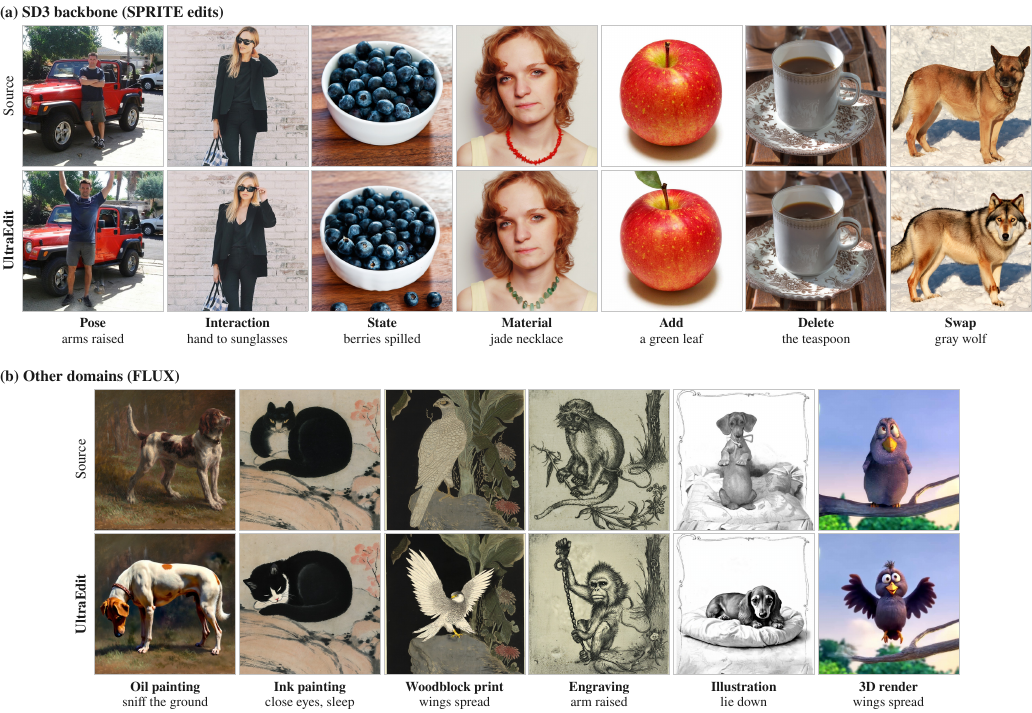}
\caption{\textbf{SD3 backbone and other domains.}
\textbf{(a)}~On Stable Diffusion~3 Medium (\cref{tab:supp-sd3}), \method makes pose, interaction, state, material, add, delete, and swap edits, while the rest of the photo stays.
\textbf{(b)}~On FLUX, from an oil painting to a 3D render, the pose changes and the medium stays.}
\label{fig:supp-apps-a}
\end{figure*}

\paragraph{SD3.}
On Stable Diffusion 3 Medium~\cite{esser2024sd3} at $512^2$, the coupling, the start times, scaling back, and the rules of the map are the code of the image editor, byte for byte; only the model calls differ.
Keys and values are shared in SD3's joint blocks $0$--$2$, without a positional weight, as SD3 has no rotary positions.
Both copies use guidance $4.5$, $16$ steps on a shift-3 grid, and $t_0=0.85$.
On $41$ \sprite edits (\cref{tab:supp-sd3}), \method passes $70.7\%$ against $56.1\%$ for the released SD3 editor ($+14.6$ points), with $72$ NFE against $132$; \cref{fig:supp-apps-a}\,(a) shows further \sprite edits.

\begin{table}[t]
\centering
\caption{\textbf{Port to SD3.} The update passes more of the edits than the released SD3 editor, with fewer NFE.
NFE counts transformer rows: two per guided call on SD3, one per call on FLUX.
Score: mean judged edit probability (\%); Edit as in \cref{sec:exp-setup}.
Bold: best within each generator; underlined: second best on SD3.
Settings: 41 \sprite edits.}
\label{tab:supp-sd3}
\small
\setlength{\tabcolsep}{2.5pt}
\resizebox{\ifdim\width>\columnwidth\columnwidth\else\width\fi}{!}{%
\begin{tabular}{llc cc}
\toprule
 & & NFE & Edit\,$\uparrow$ & Score\,$\uparrow$ \\
\midrule
SD3 & FlowEdit (released) & 132 & 56.1 & 60.4 \\
\oursvar & box, no sharing & 64 & 65.9 & 62.8 \\
\oursvar & box, sharing & 64 & \underline{68.3} & \underline{64.9} \\
\ours & \textbf{\method (Ours)}, attention, sharing & 72 & \textbf{70.7} & \textbf{65.9} \\
\midrule
FLUX & FlowEdit (released) & 48 & 58.5 & 58.8 \\
\ours & \textbf{\method (Ours)} & 38 & \textbf{78.0} & \textbf{72.1} \\
\bottomrule
\end{tabular}}
\end{table}

\begin{figure*}[tp]
\centering
\includegraphics[width=\textwidth]{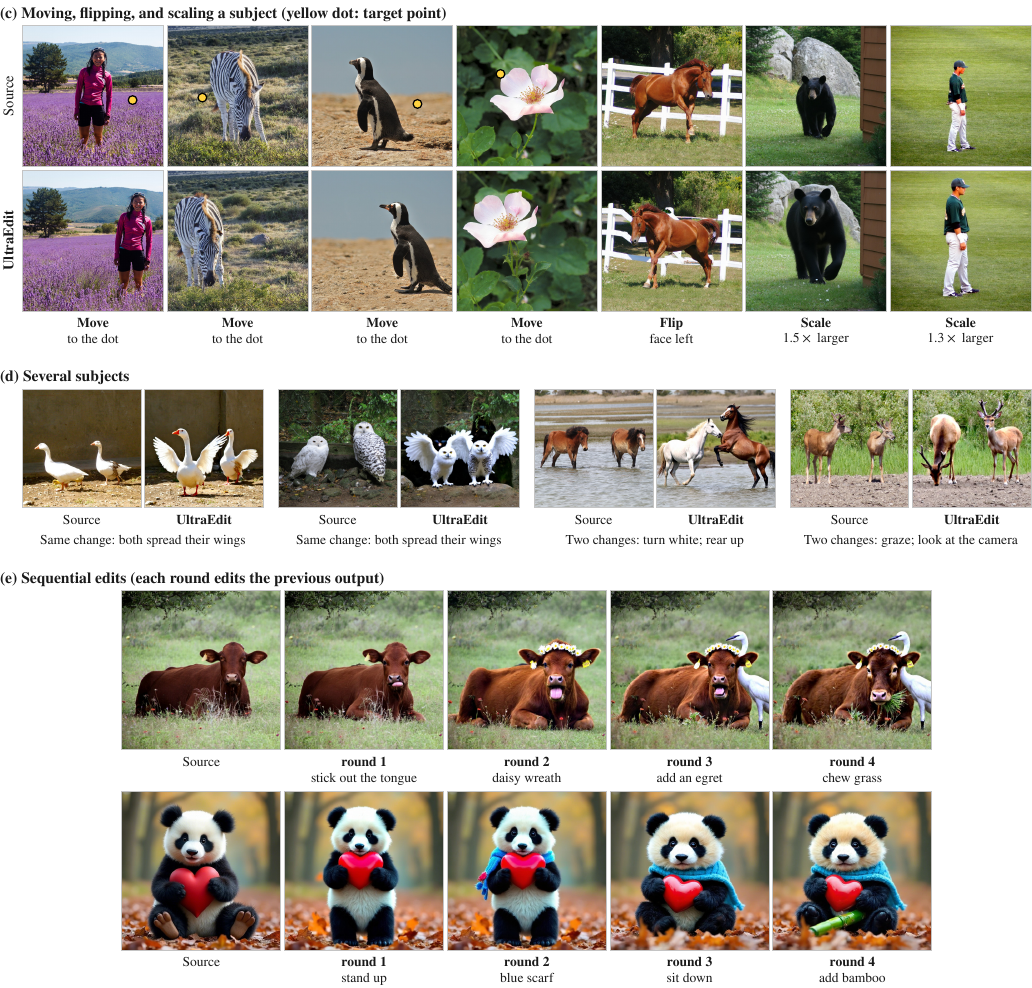}
\caption{\textbf{Moving, several subjects, and sequential edits.}
\textbf{(c)}~The forward path starts from the source with the subject moved to the yellow dot, mirrored, or scaled: the subject lands on the dot, faces the other way, or grows in place, and the place it leaves is filled with background.
\textbf{(d)}~The unchanged editor on two subjects, one change for both or one per subject.
\textbf{(e)}~Four rounds on one image, each editing the previous output; every round passes the judge.
\method on FLUX.}
\label{fig:supp-apps-b}
\end{figure*}

\paragraph{Moving to a point.}
A move is the same update on FLUX with another source for the forward path (\cref{fig:supp-apps-b}\,(c)): a map $T$ moves the subject, and the forward path is drawn from the moved source $X_w=T_\#\Xsrc$ (a flip or a scale is done in pixels, at no extra NFE).
The subject's part of $\mu$ moves with it, and the hole $N$ it leaves has no source: tokens of $\mu$ start at $t=1$, tokens of $N$ at $\tau_N=0.76$, and the others at $0.919$; on $N$ the update replaces the source estimate by $X_w$, and scaling back does not act on $N$.
With $T$ the identity and $N$ empty the code reduces bit for bit to \method.
On $37$ cases from $22$ \sprite photos (pure moves $20$, move and pose $7$, scale $5$, flip $5$) and $18$ further moves, a move succeeds if the center of the subject's OWLv2 box lies within $41$ pixels of the target point and no copy remains at the old place.
Over seeds $0$--$3$, pure moves succeed in $73$ of $80$ runs ($91\%$; $20$ cases, four seeds each), against $68$ of $80$ ($85\%$) for cut-and-paste with FLUX SDEdit inpainting of the hole; at seed $0$, prompt-only editing by FlowEdit moves the subject in none of the $20$ pure moves, and with $\tau_N=1$ pure moves succeed in $2$ of $20$.

\paragraph{Several subjects.}
\method runs unchanged on $20$ edits of $15$ Open Images photos with $43$ subjects (judged by Qwen2.5-VL-32B; \cref{fig:supp-apps-b}\,(d) shows further examples).
It edits 84\% of the subjects (36/43), against 79\% for FlowEdit; all subjects of an edit succeed on 75\% and 70\% of the edits, and the background L1 is 14.4 and 16.6.

\paragraph{Sequential edits.}
Eight chains of four edits on \sprite photos compare \method with FlowEdit at its released settings; \cref{fig:supp-apps-b}\,(e) shows two further chains.
All four rounds succeed on 5/8 chains (62.5\%), against 1/8, and after four rounds the background L1 is 23.0, against 34.8.

\paragraph{Other domains.}
On $12$ paintings, prints, illustrations, cartoons, and 3D renders under CC0, public-domain, or CC BY terms, with $20$ edits ($15$ of them pose; judged by Qwen2.5-VL-32B; \cref{fig:supp-apps-a}\,(b)), \method passes 13/20 against 9/20 for FlowEdit at its released settings, prints 75\% against 0\% and cartoons 50\% against 0\%, with a lower background L1 (11.1 against 14.9).

\paragraph{Seeds.}\phantomsection\label{supp:seeds}
On $8$ non-rigid \sprite edits, chosen among those \method passes at seed 0 so that diversity is measured on successful edits (FlowEdit passes $6$ of them at seed 0), seeds 1--7 succeed on 96.4\% (54/56), against 75.0\% for FlowEdit; because of this selection, the rates measure consistency across seeds, not edit success, which \cref{tab:main} reports.
The successful outputs differ more in the subject (DINO distance between subject crops 0.205 against 0.076; pairwise LPIPS on a box around the subjects 0.402 against 0.228), while the background L1 to the source stays within $0.5$ gray levels of FlowEdit's (17.9 against 17.4; \cref{fig:diversity}).

\paragraph{Cells of \cref{fig:teaser}.}
The sound cells use the gated editor above and the speech cells the warp variant; neither shares attention.
The door-knock and speech sources are development clips, the dog bark is generated by TangoFlux, and the knock cell labeled ``less'' removes only the third group of knocks.
Two 3D cells (a man raising his arms and a blue car) come from the first 3D run, before the three fixes above; the rabbit is one of the $127$ edits.
Six (the square top, the truck, the cowboy hat, sitting, the wings, and the metal chair) take $\mu$ from cross-attention, as the image editor does; the hat and the metal chair use reworded prompts.

\end{document}